\documentclass[preprint,12pt]{elsarticle}

\usepackage{graphicx}
\usepackage{subcaption} 
\usepackage{float}

\usepackage{algorithm}
\usepackage{algorithmic}

\usepackage{amsmath} 
\usepackage{amssymb} 

\usepackage{changepage}

\usepackage{tabularx}

\usepackage{url}

\usepackage[table,xcdraw]{xcolor} 

\usepackage{booktabs}
\usepackage{caption}

\usepackage{enumitem}

\definecolor{green}{rgb}{0.0, 0.5, 0.0}

\journal{Information Systems}

\begin{document}

\begin{frontmatter}



\title{Model-driven Stochastic Trace Clustering}

\author[kul]{Jari Peeperkorn}
\author[kul]{Johannes De Smedt}
\author[kul]{Jochen De Weerdt}

\affiliation[kul]{organization={Research Center for Information Systems Engineering (LIRIS), KU Leuven},
            city={Leuven},
            country={Belgium}}

\begin{abstract}
Process discovery algorithms automatically extract process models from event logs, but high variability often results in complex and hard-to-understand models. To mitigate this issue, trace clustering techniques group process executions into clusters, each represented by a simpler and more understandable process model. Model-driven trace clustering improves on this by assigning traces to clusters based on their conformity to cluster-specific process models. However, most existing clustering techniques rely on either no process model discovery, or non-stochastic models, neglecting the frequency or probability of activities and transitions, thereby limiting their capability to capture real-world execution dynamics. We propose a novel model-driven trace clustering method that optimizes stochastic process models within each cluster. Our approach uses entropic relevance, a stochastic conformance metric based on directly-follows probabilities, to guide trace assignment. This allows clustering decisions to consider both structural alignment with a cluster’s process model and the likelihood that a trace originates from a given stochastic process model. The method is computationally efficient, scales linearly with input size, and improves model interpretability by producing clusters with clearer control-flow patterns. Extensive experiments on public real-life datasets demonstrate that while our method yields superior stochastic coherence and graph simplicity, traditional fitness metrics reveal a trade-off, highlighting the specific utility of our approach for stochastic process analysis.
\end{abstract}



\begin{keyword}
Process Mining \sep Trace Clustering \sep Stochastic Models \sep Entropic Relevance


\end{keyword}

\end{frontmatter}


\section{Introduction}

Cluster analysis is a fundamental technique in data mining and machine learning, aimed at grouping entities such that those within the same cluster are more similar to each other than to those in different clusters. In process mining, trace clustering applies this principle to event logs by grouping process executions (cases) into smaller, more coherent subsets. This is particularly valuable for complex processes, as it enables the discovery of simpler, more interpretable process models for each cluster. Such models enhance the readability of discovered process models and can also improve the accuracy of conformance checking~\cite{DeWeerdt_2013,Greco2006,Song2009}. However, many existing trace clustering methods are unsupervised and optimize generic similarity objectives that are not explicitly aligned with improving process model quality or interpretability~\cite{DeWeerdt_2013}. Model-driven trace clustering addresses this by  iteratively discovering a process model for each cluster and assigning traces based on how well they conform to these models~\cite{DeWeerdt_2013,Ferreira2007}. 

In parallel, the emergence of the stochastic process mining paradigm has highlighted the importance of capturing the stochastic nature of real-world processes. Unlike purely structural models, stochastic process models account for the likelihood of behaviors, making them more suitable for applications such as simulation, prediction, and representative modeling of observed behavior. 

Despite these developments, existing model-driven clustering methods have yet to incorporate the stochastic characteristics of event logs. To our knowledge, no approach currently integrates stochasticity into the clustering process to produce stochastic process models per cluster.
To address this gap, we introduce a novel model-driven trace clustering method that incorporates stochastic conformance measuring into clustering decisions. Our method leverages entropic relevance (ER)~\cite{Alkhammash2022}, a stochastic conformance measure that quantifies the compression cost of traces in an event log based on directly follows probabilities. By minimizing ER within each cluster, the algorithm optimizes the representativeness of that cluster’s Directly-Follows Graph (DFG), one of the most widely used model types in (commercial) process mining tools~\cite{Vanderaalst2019}. This focus makes our method particularly advantageous in scenarios where understanding the probability of process paths is as critical as the structural flow itself.

Our approach is computationally efficient, scaling linearly with the size of the event log, and supports integration with edge frequency filtering to mitigate the “spaghetti model” problem, potentially still present in downstream clusters. By promoting clusters with shared, high-probability directly-follows relations, it enhances model clarity and interpretability.
Through extensive experiments on publicly available real-world datasets, we demonstrate that our method effectively captures control-flow behavior and yields interpretable, representative stochastic models. The results also reveal that incorporating stochasticity significantly alters the relative performance of clustering algorithms, suggesting that Entropic Clustering is the preferred choice for tasks requiring high stochastic precision, whereas traditional methods may still suffice for purely structural conformance.

To this end, our work makes the following key contributions:

\begin{itemize}
    \item We introduce stochastic conformance-based model-driven trace clustering, explicitly incorporating probabilistic behavior in clustering decisions.
    \item We propose an efficient algorithm using entropic relevance, optimizing within cluster process model coherence while maintaining scalability. 
    \item We provide a comprehensive evaluation framework, including diverse metrics to assess the impact of stochasticity on clustering rankings and process model interpretability. 
\end{itemize}

Our work is directly relevant to industry applications, particularly DFG discovery tools, which frequently rely on cluster-like representations to extract common and exceptional behaviors. Given its low computational complexity, our method provides a practical extension to existing process mining visualization and conformance checking techniques for large-scale event logs. 

The remainder of this paper is structured as follows: Section~\ref{sec:prelims} introduces key terminology and Section~\ref{sec:related} discusses relevant background literature. Section~\ref{sec:methodology} explains the workings of the entropic clustering algorithm. The experimental setup is detailed in Section~\ref{sec:setup}, followed by the results presented in Section~\ref{sec:results}. Section~\ref{sec:discussion} discusses these results along with potential limitations of the current approach. Finally, Section~\ref{sec:conclusion} provides a conclusion and outlines possible directions for future research. The code and the full results can be found online\footnote{\url{https://github.com/jaripeeperkorn/EntropicClustering}}.

\section{Preliminaries}
\label{sec:prelims}

Event logs record executions of business process activities. Each \textit{event} typically includes a case identifier (\textit{case ID}) that identifies the process instance it belongs to, an \textit{activity label} indicating which action was performed, and a \textit{timestamp} or another attribute that defines the event’s position in the sequence. The sequence of events for a case is referred to as a \textit{trace}. Let $\mathcal{A}$ be a finite set of activity labels. A trace $\sigma$ is then a finite sequence of such labels, that is, $\sigma \in \mathcal{A}^*$, and we denote its length by $|\sigma|$.
An \textit{event log} $L$ can be regarded as a multiset of traces, where each trace represents a process instance (case). Formally, $L$ is defined as a mapping $L : \mathcal{A}^* \to \mathbb{N}$, where for any $\sigma \in \mathcal{A}^*$, $L(\sigma)$ denotes the multiplicity (frequency) of $\sigma$ in the log.
In this work, we focus solely on control-flow information and represent traces as sequences of activity labels, disregarding other event attributes such as resources or timestamps. The set of all unique traces in an event log is called the set of (control-flow) \textit{variants}, with each variant's frequency denoted by its multiplicity. A \textit{variant log} is an abstraction of the full event log that captures all distinct variants and their counts.

To analyze behavior, traces are often compared with a \textit{process model} that graphically depicts the logical structure and dynamics of a business process. Process models can be manually designed or automatically discovered from event logs through \textit{process model discovery} techniques, which aim to generate models that reflect and generalize the observed behavior. \textit{Stochastic process models} extend traditional process models by incorporating probabilities (or other quantitative aspects) of process executions, thereby providing insights into the likelihood of different behaviors. While various notations exist (e.g., Petri nets~\cite{Petri}), in this work, we mainly use the \textit{Directly-Follows Graph} (DFG) as our modeling formalism. A DFG is a directed graph $G = (N, E)$, where $N \subseteq \mathcal{A}$ is the set of nodes (activities) and $E \subseteq N \times N$ is the set of edges representing directly-follows (DF) relations observed in the log. The directly-follows count (DF) between two activities $a$ and $b$, denoted $|a \to b|$, is the total number of times activity $a$ is immediately followed by $b$ in the cases. DFGs are commonly used in commercial tools to visualize the process executions found in an event log~\cite{Vanderaalst2019}.DFGs can be turned into \textit{stochastic process models} by annotating edges with transition probabilities. The transition probability is then defined as:$$P(b \mid a) = \frac{|a \to b|}{\sum_{x \in N} |a \to x|}$$This property allows us to treat DFGs as stochastic models, where the probability of a trace can be computed as the product of the transition probabilities of its sequence of activities. 
\textit{Trace clustering} refers to the task of grouping similar traces based on control-flow patterns or other attributes. Clustering can help uncover subprocesses or behavioral variants, facilitating the discovery of simpler and more interpretable models for each group.

\section{Related work}
\label{sec:related}
\subsection{Process discovery \& conformance checking}
Well-known methods for automatically discovering Petri Nets include Heuristic Miner~\cite{Weijters2011} and Inductive Miner~\cite{Leemans2014}, while other methods focus on other model types, such as Fuzzy Miner~\cite{Gunther2007} that can be used to create graph representations of the process at different levels of detail. \textit{Conformance checking} techniques are used to compare an event log with a given process model, along different dimensions. Fitness measures how well the behavior found in the event log is described by the model, while precision measures how much of the allowed behavior is actually present in the event log. Typical fitness and precision metrics include token-replay~\cite{rozinat2008} and alignments~\cite{Vanderaalst2012,Adriansyah2015}. Generalization metrics aim to measure the model's usefulness on unseen (but correct) process behavior. Simplicity reflects the readability, understandability, and complexity of the process, measured, for example, by the number of arcs~\cite{SanchezGonzalez2010}. \textit{Stochastic conformance checking} techniques incorporate the  probabilities found in stochastic process models, e.g., by analyzing the statistical properties of execution patterns~\cite{Leemans2019,Leemans2020,Leemans2023}. 

\subsection{Trace clustering}
A common line of trace clustering work relies on \textit{vector-based} methods, which transform traces into numerical feature vectors and apply standard clustering algorithms~\cite{Greco2006,Song2009,DeKoninck2018}. \textit{Context-aware} approaches extend this by incorporating additional information about the execution context or structural dependencies~\cite{Bose2010}. A more process-centric alternative is provided by \textit{model-based} clustering techniques. Ferreira et al.~\cite{Ferreira2007} proposed a method that learns a mixture of first-order Markov models using the Expectation-Maximization algorithm, but its computational cost limits scalability. ActiTraC~\cite{DeWeerdt_2013} is another model-based approach that iteratively mines models using the Heuristic Miner algorithm~\cite{Weijters2011} and evaluates the assignment of traces to clusters using fitness scores. Other works make use of alignments to perform trace clustering in different ways as well: using local alignments in combination with k-means clustering~\cite{Evermann2016}, by using model traces as centroids~\cite{Chatain2017}, or by explicitly using an available process model~\cite{Boltenhagen2019}.
Despite existing advances, no method has yet integrated stochastic conformance information into the clustering process. In particular, model-driven clustering approaches have not leveraged the probabilistic behavior captured by stochastic process models. This gap motivates our contribution: a trace clustering technique that uses entropic relevance to evaluate and guide trace assignments, offering a distinct alternative that prioritizes stochastic coherence over purely structural alignment.

\section{Model-driven stochastic clustering}
\label{sec:methodology}
This section presents the foundations and implementation of our proposed approach, including the entropic relevance metric, the entropic clustering algorithm, and the initialization strategies used.

\subsection{Entropic Relevance}
We use entropic relevance (or ER), a stochastic conformance checking measure that quantifies the information-theoretic compression cost of traces in an event log, given the behaviour of a stochastic process model as induced by its structure and transition probabilities~\cite{Alkhammash2022}. A lower compression cost (i.e., a lower ER score) indicates a better fit between the model and the event log. For each trace, the ER contribution is computed using the probability $p$ assigned to that trace by the model, calculated as: $- \text{log}_2(p)$. The total ER score for a given log-model combination is obtained by summing these contributions over all traces in the log. For non-fitting traces, a so-called background model is used, giving a compression cost that is completely based on the log only (and is generally much higher). ER gives a score that measures quality along both the fitness and precision dimensions. By punishing for a lack of fitness through the background model, or from a stochastic point of view, lower values for $p$ are assigned to highly unlikely cases. If the model allows too much behavior (with high probability), i.e., it is imprecise, the probabilities of the fitting cases will drop as well, leading to an increase in ER score.  

In our approach, we use DFGs, which can be converted to stochastic process models. The discovery of a DFG can be done fast and efficiently, by just counting how often each activity appears and the frequency with which one activity directly follows another in the traces of the event log. For our purposes, the DFG is stored as two dictionaries: one containing the activity types (nodes) and their counts, and one the directly-follows relations (edges) and their counts. Given a DFG, the probability of a specific trace is computed as the product of the transition (edge) probabilities along the trace. Each transition probability is calculated by dividing the count of the corresponding edge by the count of its source node. In the original ER implementation, stochastic process models are converted into stochastic deterministic finite automata (SDFAs) to compute trace probabilities. To replicate this in the DFG-based approach, special Beginning-Of-Sequence (BOS) and End-Of-Sequence (EOS) activities are introduced to all traces and included in the DFG. Incorporating these artificial transitions effectively replicates the initialization and termination probabilities that would otherwise be handled in an SDFA. The traces with the added BOS and EOS activities are also the ones used to discover the DFG, i.e., the DFGs will also include those nodes. Since all traces now start with BOS and end with EOS, the transition probabilities from the BOS node to any other node can be seen as initialization probabilities, and those from each node to EOS as termination probabilities.

\begin{algorithm}
    \algsetup{linenosize=\tiny}
    \scriptsize
    \captionsetup{font=footnotesize}
    \caption{Compute (simplified) Entropic Relevance (\texttt{ER}) for a DFG-based Model}
    \begin{algorithmic}[1]
        \REQUIRE Variant log \(L = \{(t_i, m_i)\}_{i=1}^K\):  $K$ distinct variants $t_i$, each with multiplicity $m_i$ \\
        directly-follows graph \(G=(N,E,c_N,c_E)\): with \\
        $N =$ set of activity labels incl. BOS and EOS\\ 
        $E =$ set of directed edges \\
        $c_N$ and $c_E$ the count of occurrences of the nodes and edges (the DFG). \\
        underflow threshold \(\varepsilon\)
        \ENSURE  Average \(\mathrm{ER}\) score
        
        \STATE $T \gets \sum_{i=1}^K m_i$  \COMMENT{Total number of trace occurrences}
        \STATE $\mathrm{ER}_{\text{sum}} \gets 0$  \COMMENT{Sum of per-trace information costs}

        \FORALL{$(\alpha,\beta)\in E$}
            \STATE $P(\beta\mid\alpha) \gets c_E(\alpha,\beta) / c_N(\alpha)$
            \COMMENT{Probability of transition $\alpha\to\beta$}
        \ENDFOR

        \FOR{$i = 1$ \TO $K$}
            \STATE Let $t_i = \langle a_0, a_1, \dots, a_\ell\rangle$  \COMMENT{$a_0=\mathrm{BOS},\,a_\ell=\mathrm{EOS}$}
            \STATE $p \gets 1$  \COMMENT{Initialize probability of $t_i$}

            \FOR{$j = 1$ \TO $\ell$}
                \STATE $p \gets p \times P(a_j\mid a_{j-1})$
            \ENDFOR

            \STATE $\hat p \gets \max(p, \varepsilon)$  \COMMENT{Ensure $\hat p \ge \varepsilon$}

            \STATE $\mathrm{ER}_{\text{sum}} \gets \mathrm{ER}_{\text{sum}} - m_i\,\log_2(\hat p)$
            \COMMENT{$m_i$: multiplicity of $t_i$}
        \ENDFOR

        \STATE $\overline{\mathrm{ER}} \gets \mathrm{ER}_{\text{sum}} / T$  \COMMENT{Optional: Divide by total traces}
        \RETURN $\overline{\mathrm{ER}}$
    \end{algorithmic}
    \label{alg:ER}
\end{algorithm}

Our (simplified) ER calculation is formalized in Algorithm~\ref{alg:ER}. The procedure begins by calculating the total number of traces $T$ based on the multiplicities of the variants (Line 1) and initializing the total ER accumulator (Line 2). Next, the algorithm precomputes the transition probability for every edge in the DFG (Lines 3–5); this is done by dividing the count of a specific edge by the count of its source node. The algorithm then iterates through every distinct trace variant $t_i$ in the log (Line 6). As noted above, each trace is treated as a sequence starting with BOS and ending with EOS (Line 7). We initialize the probability $p$ for the current trace (Line 8) and calculate its total probability by multiplying the transition probabilities of every consecutive pair of activities along the path (Lines 9–11). To handle numerical instability, we safeguard the calculated probability against underflow by ensuring it is at least $\varepsilon$ (Line 12). We then calculate the information cost (ER) and add it to the running sum (Line 13). 

Usually, ER of a log-model combination is calculated as the total ER (over all traces)~\cite{Alkhammash2022}. To be able to be able to compare ER scores across different clusters (or event logs) with different amounts of traces, we also opted to use the average ER score over traces (Line 15). In the clustering algorithms, we calculate the ER value of individual traces only, so there is no difference. However, in the full results (see \ref{sec:FullResults}), we report both the average ER and the full sum. 

Additionally, as mentioned above, the original ER formulation also handles non-fitting traces using a background model. However, in our method, ER scores are computed only for trace–DFG combinations in which the trace was used to discover the unfiltered DFG. The model therefore always fits the trace, and its probability can be directly derived from the DFG without resorting to a background model. Finally, as mentioned before, to avoid numerical underflow when computing probabilities for complex models (e.g., traces with many loops), our implementation enforces a lower bound on trace probabilities. Specifically, each probability is constrained to be at least $\varepsilon = 10^{-10}$ before the logarithmic transformation. This effectively introduces a ceiling on the information cost for traces with extremely low likelihood. While this may yield slightly optimistic absolute ER scores for very complex (sub)processes, we argue that for the purpose of clustering, where the focus lies on the interpretability of the process models in each cluster, distinguishing between ``extremely unlikely'' and ``infinitesimally unlikely'' traces provides little additional value.

\subsection{Entropic Clustering}

\begin{figure}
    \centering
    \includegraphics[width=0.9\linewidth]{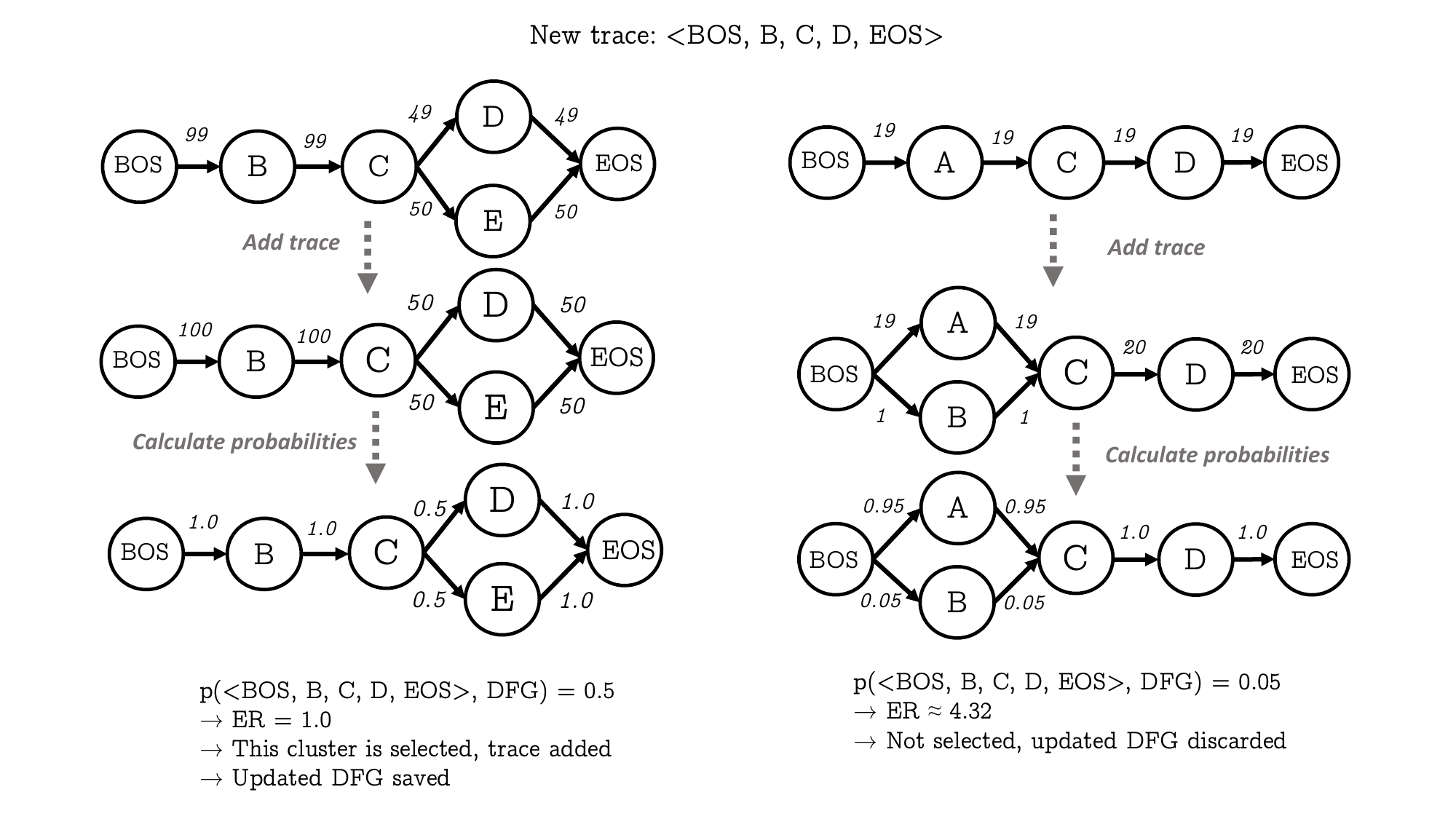}
    \caption{Illustrative example of how the method selects the cluster.}
    \label{fig:example_clustering}
\end{figure}

Our method, called Entropic Clustering (EC), aims to minimize the entropic relevance (ER) scores of the cluster-DFG pairs (i.e. the ER score of the cluster of the DFG obtained from that cluster) within each cluster, as outlined in Algorithm~\ref{alg:EC}. To improve efficiency, we make use of a variant log, consisting of the unique control-flow variants and their occurrence counts, rather than processing each individual case separately. The algorithm begins by specifying the required number of clusters, $k$, and selecting initial seed variants (explained below) to form the first clusters. A DFG is then discovered for each cluster. Next, we iterate through the remaining variants in order of prevalence, compute their ER scores with each cluster’s DFG, and assign them to the cluster where they achieve the lowest ER score. After assignment, that cluster's DFG is updated by adjusting its node and edge counts accordingly. 
This assignment process is illustrated in Figure~\ref{fig:example_clustering}. The trace is hypothetically added to each cluster’s DFG, its ER score is computed, and it is assigned to the cluster where the ER increase is minimal, optimizing the DFG graph to have as few low frequency DFs as possible.

Since ER computation scales linearly with input size \cite{Alkhammash2022}, the clustering process (excluding initialization) is also linear in complexity with the exact runtime depending on trace lengths, the preselected number of clusters, and the number of unique variants. An alternative approach, which adds the variant to the cluster for which the total cluster ER score is minimal (after adding), rather than the specific variant's ER, was also explored. However, initial results showed inferior performance, leading us to omit it from further experiments.

        
        

\begin{algorithm}
    \algsetup{linenosize=\tiny}
    \scriptsize
    \captionsetup{font=footnotesize}
    \caption{Entropic Clustering (\texttt{EC})}
    \begin{algorithmic}[1]
        \REQUIRE Event log \( L = \{(t_i, m_i)\}_{i=1}^K \), where \( t_i \) is a trace variant and \( m_i \) its multiplicity \\
        number of clusters \( k \) \\
        underflow threshold \(\varepsilon\)
        
        \ENSURE Partitioning of \( L \) into \( k \) clusters \( \mathcal{C} = \{C_1, \ldots, C_k\} \)

        \STATE Let \( \mathcal{V} \gets \{t_1, \ldots, t_K\} \) be the set of unique variants in \( L \)
        \STATE Select \( k \) initial seeds \( \{s_1, \ldots, s_k\} \subseteq \mathcal{V} \) using a chosen initialization method
        \FOR{$j = 1$ to $k$}
            \STATE Initialize cluster \( C_j \gets \{s_j\} \)
            \STATE Initialize DFG \( G_j = (N, E, c_N, c_E) \) from seed \( s_j \):
            \STATE \quad \( N \gets \) Activity labels in \( L \cup \{\mathrm{BOS}, \mathrm{EOS}\} \)
            \STATE \quad \( E \gets \) Directed edges observed in \( L \)
            \STATE \quad \( c_N, c_E \gets \) Counts of nodes/edges in \( s_j \) (starting counts)
        \ENDFOR

        \FOR{each variant \( v \in \mathcal{V} \setminus \{s_1, \ldots, s_k\} \), in decreasing order of multiplicity}
            \STATE Let \( L_{temp} \gets \{(v, 1)\} \) \COMMENT{Prepare single-instance log for ER calc}
            \FOR{each cluster \( C_j \in \mathcal{C} \)}
                \STATE \textbf{Temporarily update} \( G_j \) for variant \( v \):
                \STATE \quad Increment \( c_N \) and \( c_E \) by 1 for all nodes/edges in path of \( v \)
                
                \STATE \( \text{score}_j \gets \textsc{ComputeER}(L_{temp}, G_j, \varepsilon) \)
                \COMMENT{Call Algorithm \ref{alg:ER}}
                \STATE \textbf{Revert} changes to \( G_j \) (subtract 1 from relevant counts)
            \ENDFOR
            \STATE Let \( j^* = \arg\min_{j} \text{score}_j \)
            \STATE Assign \( v \) to cluster \( C_{j^*} \)
            \STATE Update \( G_{j^*} \) permanently (add actual multiplicity \( m_v \) to counts)
        \ENDFOR

        \RETURN Final cluster assignment \( \mathcal{C} = \{C_1, \ldots, C_k\} \)
    \end{algorithmic}
    \label{alg:EC}
\end{algorithm}

In addition to the standard algorithm, we also investigate a recursive, hierarchical variant of Entropic Clustering, denoted as \texttt{EC\textsubscript{split}}. This approach is detailed in Algorithm~\ref{alg:ECsplit}. Instead of initializing $k$ clusters simultaneously, \texttt{EC\textsubscript{split}} begins with a single cluster containing the entire event log. It then proceeds iteratively: in each step, the algorithm identifies the ``worst-performing'' cluster—defined as the cluster with the highest total (or average) Entropic Relevance score. This high-entropy cluster is then split into two sub-clusters using the standard Entropic Clustering procedure with $k=2$. This process repeats until the pre-defined total number of clusters is reached. The theoretical motivation for this greedy strategy is drawn from recursive partitioning methods like ID3 for decision trees~\cite{Quinlan1986}. The hypothesis is that by focusing computational effort on refining the most unstructured subsets of traces (those with the poorest model fit), the algorithm might converge to a better global solution or achieve comparable results with lower computational overhead. We evaluate this hypothesis against the standard global initialization in our experimental section. 


\begin{algorithm}
\captionsetup{font=footnotesize}
\algsetup{linenosize=\tiny}
\scriptsize
\caption{Entropic Clustering Split (\texttt{EC\textsubscript{split}})}
\begin{algorithmic}[1]
    \REQUIRE Event log \( L = \{(t_i, m_i)\}_{i=1}^K \) \\
    desired number of clusters \( k \) \\
    underflow threshold \(\varepsilon\)
    \ENSURE Partitioning of \( L \) into \( k \) clusters \( \mathcal{C} = \{C_1, \ldots, C_k\} \)

    \STATE Initialize clustering \( \mathcal{C} \gets \textsc{EC}(L, 2, \varepsilon) \) 
    \COMMENT{Call Entropic Clustering (Alg. \ref{alg:EC})}
    
    \WHILE{\( |\mathcal{C}| < k \)}
        \FOR{each cluster \( C_j \in \mathcal{C} \)}
                \STATE Let \( L_j \subseteq L \) be the sub-log containing variants assigned to \( C_j \)
                \STATE Build DFG \( G_j \) from \( L_j \) (calculating \( N, E, c_N, c_E \))
                \STATE \( \text{score}_j \gets \textsc{ComputeER}(L_j, G_j, \varepsilon) \)
                \COMMENT{Calculate ER (Alg. \ref{alg:ER})}
            \ENDFOR

        \STATE Let \( j^* = \arg\max_j \text{score}_j \) 
        \COMMENT{Identify cluster with highest entropy}
        \STATE Let \( L_{j^*} \) be the variant log of cluster \( C_{j^*} \)
        \STATE \( \{C_{new1}, C_{new2}\} \gets \textsc{EC}(L_{worst}, 2, \varepsilon) \)
            \COMMENT{Split the worst cluster}
        \STATE Update \( \mathcal{C} \gets (\mathcal{C} \setminus \{C_{j^*}\}) \cup \{C_{new1}, C_{new2}\} \)
    \ENDWHILE

    \RETURN Final cluster assignment \( \mathcal{C} \)
\end{algorithmic}
\label{alg:ECsplit}
\end{algorithm}

\subsection{Initialization}

The EC algorithm requires initial seed selection, which determines the first traces assigned to each cluster. The simplest approach is random selection, but since these initial seeds remain in their respective clusters, their selection can significantly impact clustering quality. To address this, we propose a more sophisticated \texttt{++} initialization (Algorithm~\ref{alg:++}), inspired by how \texttt{k-means++} clustering handles initialization~\cite{Arthur2007}. This method enhances seed selection by considering pairwise ER distances between traces. This initialization starts again by picking the first seed at random. Then, the \texttt{pairwise ER distance} between this seed and all other variants (traces) is computed. The pairwise ER is calculated by constructing a DFG from just these two traces, and calculating the average ER scores over both traces. The more similar those two traces are, the lower this score (analogous to edit distance). The next seed is selected randomly, weighted by the square of its pairwise ER distance from the current seed. We then continue similarly for the next seeds (if needed), but now calculate the pairwise ER with each seed for each variant (trace). For each variant, we use the minimum distance, i.e., the distance to the closest seed, as weights when selecting the next seed. We continue until we have a number of seeds equal to our required number of clusters.  

    

\begin{algorithm}
\captionsetup{font=footnotesize}
\algsetup{linenosize=\tiny}
\scriptsize
\caption{\texttt{++} Initialization for Entropic Clustering}
\begin{algorithmic}[1]
    \REQUIRE Set of trace variants \( \mathcal{V} = \{t_1, \ldots, t_K\} \), desired number of seeds \( k \)
    \ENSURE Set of selected seeds \( \mathcal{S} = \{s_1, \ldots, s_k\} \subseteq \mathcal{V} \)

    \STATE Randomly select \( s_1 \in \mathcal{V} \)
    \STATE Initialize seed set \( \mathcal{S} \gets \{s_1\} \)

    \WHILE{\( |\mathcal{S}| < k \)}
        \FOR{each variant \( v \in \mathcal{V} \setminus \mathcal{S} \)}
            \FOR{each seed \( s \in \mathcal{S} \)}
                \STATE Construct DFG from \( \{v, s\} \)
                \STATE Compute average entropic relevance \( \mathrm{ER}(v, s) \)
            \ENDFOR
            \STATE Let \( d(v) = \min_{s \in \mathcal{S}} \mathrm{ER}(v, s) \)
        \ENDFOR

        \STATE Sample new seed \( s^* \in \mathcal{V} \setminus \mathcal{S} \) with probability proportional to \( d(v)^2 \)
        \STATE Update \( \mathcal{S} \gets \mathcal{S} \cup \{s^*\} \)
    \ENDWHILE

    \RETURN Final seed set \( \mathcal{S} \)
\end{algorithmic}
\label{alg:++}
\end{algorithm}

\begin{figure}
    \centering
    \includegraphics[width=0.75\linewidth]{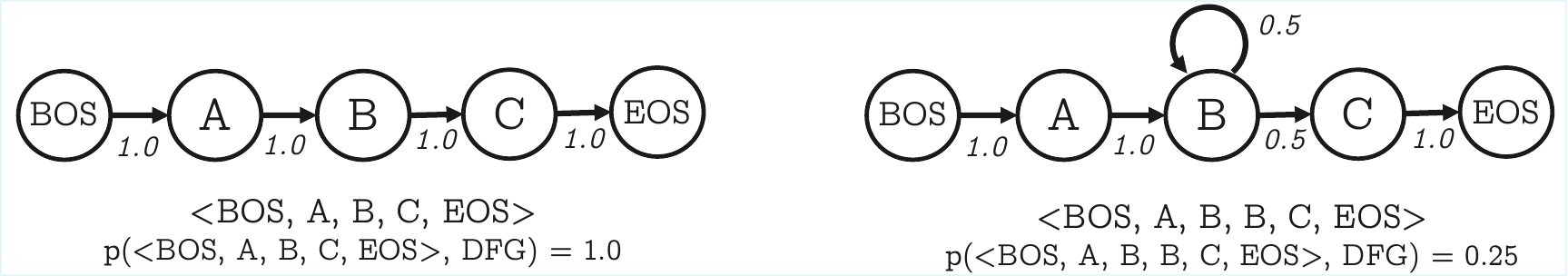}
    \caption{Illustrative example showing how loops can make the ER of a single-trace DFG non-zero.}
    \label{fig:example_init}
\end{figure}

In the absence of any kind of looping behavior, the ER of a trace with its single-trace DFG, i.e. a DFG discovered by only itself, will be equal to 0. In this sense we mean any type of repetition of activity labels looping behavior. Without any form of repetition, the single-trace DFG will of course always simply be linear, with each edge probability equal to 1. However, when we have loops this is not necessarily the case, since at least one node (the repeated activity type) will have multiple outgoing edges, leading to edge probabilities not equal to 1. If a loop is taken multiple times, the (pairwise) ER score of that trace will be high, with any DFG. This is illustrated with an example in Figure~\ref{fig:example_init}. This effect can also have a significant impact on our initialization, cases containing loops will more likely be selected as initial seeds. To mitigate this, we also implemented a normalized version \texttt{++norm} where we divide the components of the pairwise ER scores by the ER of that variant with its own single-trace  DFG. This ensures that initialization is less biased towards selecting only traces with multiple loops or other types of repeating behavior, as seeds.

\section{Experimental setup}
\label{sec:setup}

Our experimental setup evaluates the EC algorithm against various clustering methods by addressing the following key questions:
\begin{itemize}[noitemsep,topsep=0pt]
    \item \textbf{RQ1:} How uniformly structured are the clusters from a stochastic perspective?
    \item \textbf{RQ2:} How simple and readable are the discovered process models for each cluster?
    \item \textbf{RQ3:} What impact does the clustering method have on (non-stochastic) fitness and precision.
\end{itemize}

\subsection{Event logs}

In order to test our methods on a variety of different process behavior, we used 8 real-life event logs: the helpdesk event log~\cite{Helpdesk}, the road traffic fine management (RTFM) event log~\cite{RTFM}, the BPIC 2013 challenge event logs (the incidents and closed problems)~\cite{BPIC13}, the hospital billing event log~\cite{hospitalbilling}, the sepsis event log~\cite{Sepsis}, and BPIC 2012~\cite{BPIC12} and 2015 event logs~\cite{BPIC15}. Table~\ref{tab:log_info} provides an overview of the key characteristics of these event logs. 
Note that for \texttt{BPIC15} log we use a version where all 5 municipality logs are merged into one, as done before in other works concerning clustering~\cite{DeKoninck2018}.

\renewcommand{\thefootnote}{\alph{footnote}}
\setcounter{footnote}{0}
\renewcommand{\footnoterule}{}
\begin{table}[h]
\centering
\caption{Log Information}
\label{tab:log_info}
\begin{minipage}{\textwidth} 
\begin{tabularx}{\textwidth}{l*{7}{>{\centering\arraybackslash}X}}
\toprule
Log Name & \resizebox{0.10\textwidth}{!}{Total Cases} & \resizebox{0.10\textwidth}{!}{\# Variants} & \resizebox{0.09\textwidth}{!}{Voc. Size} & \resizebox{0.10\textwidth}{!}{Avg. Case Len.} & \resizebox{0.10\textwidth}{!}{Min. Case Len.} & \resizebox{0.10\textwidth}{!}{Max. Case Len.} & \resizebox{0.015\textwidth}{!}{k} \\
\midrule
\texttt{Helpdesk} & 4,580 & 226 & 14 & 4.66 & 2 & 15 & 4\\

\texttt{RTFM} & 150,370 & 231 & 11 & 3.73 & 2 & 20 & 4 \\

\texttt{BPIC13 In.\footnotemark[1]} & 7,554 & 1,511 & 4 & 8.68 & 1 & 123 & 5 \\

\texttt{BPIC13 Cl.\footnotemark[1]} & 1,487 & 183 & 4 & 4.48 & 1 & 35 & 5 \\

\texttt{Hospital Billing} & 100,000 & 1,020 & 18 & 4.51 & 1 & 217 & 5 \\

\texttt{Sepsis} & 1,050 & 846 & 16 & 14.49 & 3 & 185 & 6 \\

\texttt{BPIC12} & 13,087 & 4,366 & 24 & 20.04 & 3 & 175 & 4\\

\texttt{BPIC15} & 5,647 & 2,501 & 29 & 16.37 & 3 & 76 & 4 \\
\bottomrule
\end{tabularx}
\vspace{-0.3cm}
\footnotetext[1]{\scriptsize{By using the default importer of \texttt{pm4py} activity types were not separated into lifecycle transitions.}}
\end{minipage}
\end{table}

\renewcommand{\thefootnote}{\arabic{footnote}}

\subsection{Evaluation metrics}

Evaluating clustering methods is inherently challenging because the quality of clusters depends on the downstream task. To address this, and to stay close to the end-goal of most trace clustering pipelines in process mining, we employ multiple evaluation metrics related to the quality and readability of the discovered process models in each cluster. To align our evaluation with the typical end-goal of trace clustering in process mining, we use a diverse set of metrics that jointly assess three aspects: stochastic structure, model simplicity, and non-stochastic conformance. These aspects correspond to our research questions RQ1-RQ3 and are reflected in the metrics described below.

First, to evaluate non-stochastic conformance and model quality within each cluster, we use the Inductive Miner infrequent variant~\cite{Leemans2014} with its default noise threshold of 0.2 to discover a Petri net for each cluster. Using the clustered event log and the corresponding Petri net, we then compute several different (\textit{non-stochastic}) conformance checking metrics: namely, token-based fitness and precision~\cite{rozinat2008}, and alignment-based fitness~\cite{Vanderaalst2012} and precision~\cite{Adriansyah2015}. These metrics quantify how well the discovered model represents the behavior in the cluster and how much the clustering itself can improve these conformance measures when using the same discovery technique, which directly addresses RQ3. To capture model simplicity, we calculate the inverse arc-degree of the discovered Petri nets ~\cite{SanchezGonzalez2010}, which serves as a structural simplicity measure and addresses RQ2. For these metrics, we used the implementation available in the Python package \texttt{pm4py} (Version 2.7.8.4) ~\cite{Berti2023}.

Second, to evaluate how uniformly structured the clusters are from a stochastic perspective (RQ1), we discover an unfiltered directly-follows graph (DFG) for each cluster and compute the entropic relevance (ER)~\cite{Alkhammash2022}, following Algorithm~\ref{alg:ER}. We use a simplified ER variant that is appropriate in our setting because the discovered process models are fully fitting. ER acts as a stochastic conformance measure: high ER values indicate greater structural homogeneity within the cluster. We report both the total ER of the log-DFG combination (\texttt{ER\textsubscript{sum}}), as proposed in~\cite{Alkhammash2022}, and the average ER score over all traces in the event log (\texttt{ER\textsubscript{av}}).

To quantify the structural uncertainty of the DFG, we also calculate the total graph entropy (\texttt{graph\textsubscript{entr}}) directly on the DFG, again addressing RQ1. Specifically, we calculate the Shannon entropy of edge transitions based on the probabilities of moving from one node (activity) to another. Higher entropy indicates a more unpredictable structure, while lower entropy suggests more deterministic and structured connectivity patterns~\cite{Dehmer2011}.
\begin{equation*}
    \texttt{graph\textsubscript{entr}} = \sum_{u \in V} H_u = -\sum_{u \in V} \sum_{v \in V, p(u \to v)>0} p(u \to v) \log_2 p(u \to v)
\end{equation*}
Finally, to assess an additional aspect of model simplicity and readability (RQ2), we calculate the graph density (\texttt{graph\textsubscript{dens}}) of each DFG. Graph density is obtained by dividing the number of observed edges in the DFG by the maximum possible number of edges between activity types:
\begin{equation*}
    \texttt{graph\textsubscript{dens}} = \frac{|\text{Edges}|}{|\text{Activity types}| \cdot (|\text{Activity types}| - 1)}
\end{equation*}
\texttt{Graph\textsubscript{dens}} quantifies how close the DFG is to being fully connected and provides a complementary \textit{non-stochastic} simplicity measure that focuses on the DFG’s connectivity patterns.

\begin{figure}
    \centering
    \includegraphics[width=0.90\linewidth]{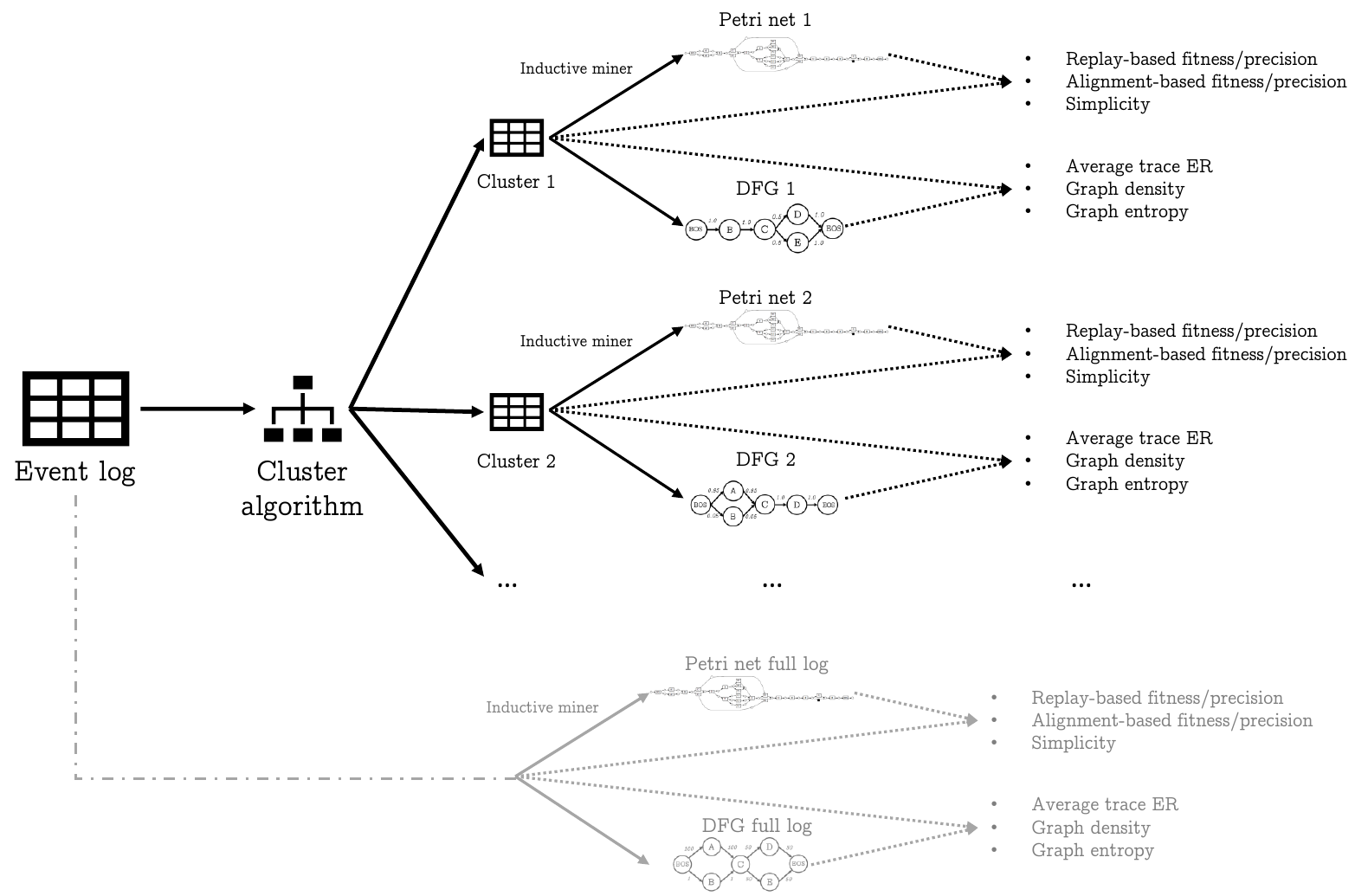}
    \caption{High-level overview of the experimental setup.}
    \label{fig:overview}
\end{figure}

A high-level overview of the experimental setup can be found in Figure~\ref{fig:overview}. 

\subsection{Baseline methods}

To assess the performance of our proposed methods, we compare them with several baseline and existing clustering approaches. First, we include two simple baselines. The first baseline does not apply any clustering, meaning that the entire event log is used when calculating the evaluation metrics. The second baseline applies a random clustering approach, where traces are randomly assigned to clusters. Beyond these baselines, we also compare our methods with a frequency-based clustering approach. In this method, we construct a frequency encoded vector for each trace, where each dimension of the vector corresponds to an activity type and the values represent the frequency of occurrences of these activities. The vectors are then normalized by dividing each value by the total number of activities in the respective trace. Once the vectors are prepared, clustering is performed using the \texttt{k-means++} algorithm~\cite{Arthur2007}. We also compare our methods to an embedding-based clustering method using \texttt{trace2vec}~\cite{DeKoninck2018}. Trace2vec is a self-supervised doc2vec-style approach~\cite{Le2014}, where trace embeddings are learned by training a shallow neural network to predict the context of activities within a sequence. In our implementation, we use a distributed bag-of-words (PV-DBOW) model trained for 200 epochs. The embedding dimension is set to the square root of the number of unique activity types, while the window size is set to two activities. We use the \texttt{Gensim} library (Version 4.3.2)~\cite{gensim}. After obtaining the trace embeddings, we again apply the \texttt{k-means++}  algorithm to obtain clusters. We further compare our approach with the active trace clustering method ActiTraC, in both its frequency-based and distance-based variant. For both ActiTraC methods we use the implementation within the \texttt{ProM} framework (Version 6.13)~\cite{ProM}, and the default parameter settings, where the minimal cluster size is set to 0.25, the target ICS-fitness is set to 1.0, and the heuristic net discovery is used. To maintain consistency across all methods, the number of clusters is set to be the same as in the other approaches. Any traces that are not initially assigned to clusters are distributed among the existing clusters.

\subsection{Elbow experiment}

Since entropic clustering (like many other trace clustering methods) requires a predefined number of clusters, we first performed an independent analysis to determine a suitable cluster count $k$ for each event log. In this preliminary parameter determination phase, we conducted an ``elbow experiment'' by plotting the values of entropic relevance, graph density, and graph entropy across a range of cluster counts (from $k=1$ to $k=10$). These metrics were selected based on their computational efficiency. We analyzed the trends for random clustering, frequency-based clustering, \texttt{EC}, and \texttt{EC\textsubscript{split}} (both with \texttt{++} initialization), also selected on their computational efficiency. 

The detailed plots from this analysis are presented in \ref{sec:Elbow}. While some event logs exhibited a distinct elbow point with consensus across metrics, others required a more qualitative selection. Crucially, this selection was made prior to the comparative evaluation.  Once a specific $k$ was selected for an event log (as listed in Table~\ref{tab:log_info}), it was fixed and applied to all clustering methods in the subsequent experiments. This ensures that our downstream evaluation compares different clustering methods on a level playing field: any potential arbitrariness in the choice of $k$ affects all methods equally and thus does not bias the relative ranking of the algorithms.

\subsection{Ranking}

To evaluate the effectiveness of each clustering method, we determine its ranking for each event log along each metric. For each metric, we then calculate the average rank of each clustering method. The statistical significance of the ranking is determined following the well-established methodology described by Dem{\v{s}}ar et al.~\cite{Demsar2006}. Specifically, for each metric, we perform Friedman’s test to assess whether the ranking of clustering methods is statistically significant across datasets. Friedman's test is a non-parametric test that uses the average ranks of clustering methods across datasets. The null hypothesis assumes that all methods perform equivalently on average (i.e., their ranks should be equal). The Friedman test statistic follows a chi-square distribution with \( \chi^2_F \) and \( k-1 \) degrees of freedom, where \( k \) is the number of clustering methods. If \( p < 0.05 \), we reject the null hypothesis, indicating that at least one method differs significantly.

Because we run a separate Friedman test for each evaluation metric (multiple tests), we control the family-wise Type I error across metrics by applying the Holm step-down correction to the Friedman $p$-values.~\cite{Holm1979}. Holm’s procedure works by sorting the $p$-values and multiplying each by $(m - \text{rank} + 1)$ where $m$ indicates the total number of metrics (8 in this case). It also enforces monotonicity: adjusted $p$-values must be non-decreasing, meaning that each corrected value is at least as large as those for all lower-ranked tests. This correction is somewhat conservative: one could reasonably argue that the metrics do not form a single fully coherent family, as some are related while others (e.g., fitness versus precision) measure very different aspects. Moreover, such corrections are most relevant when a claim hinges on any individual null hypothesis being rejected, i.e., if you treat each metric as a qualitatively distinct evaluation dimension. In our case, we are more interested in how methods compare on each metric individually than in whether a method is superior on at least one metric. Nevertheless, since some of our metrics do correlate to a certain extent, we apply Holm’s correction for completeness.

If the Holm-adjusted p-value is below 0.05, we reject the null hypothesis for that metric, indicating that at least one method differs significantly. We then proceed with post-hoc pairwise comparisons using the Nemenyi test, which determines whether the performance differences between clustering methods are statistically meaningful, and accounts for multiple comparisons. The results from the Nemenyi test are further visualized using a Critical Difference (CD) diagram, which provides an intuitive way to interpret the findings. In a CD diagram, clustering methods are placed along a ranking scale, and those that are not significantly different are connected with a horizontal line.

\section{Experimental results}
\label{sec:results}

For each event log, the same number of clusters is extracted using each clustering method. The different metrics, as described above, were computed for each cluster, alongside the metric values for the full event log, which served as a baseline. The final score for each method is determined by taking the weighted average of the scores, where the weight is based on the number of cases in each cluster, ensuring that each case is weighted equally. The detailed results for each event log can be found in ~\ref{sec:FullResults}. Due to the experiments running \textit{out of time}, the alignments-based conformance checking methods were not used for logs \texttt{BPIC12} and \texttt{BPIC15}. The full results, for each separate cluster, can be found online. Table~\ref{tab:average_rankings} presents the average rankings of each clustering method across the eight event logs, evaluated separately for each metric. The table uses a color-coded ranking system, where green represents the highest rank (1) and white represents the lowest (12). Since the rankings for \texttt{ER\textsubscript{av}} and \texttt{ER\textsubscript{sum}} are the same, we only add \texttt{ER} once. The best-performing method for each metric is highlighted in bold with an underline. The Friedman $\chi^2$ statistic and both the raw and Holm-adjusted $p$-values indicate whether at least one method significantly outperforms the others for a given metric. Statistically significant p-values (i.e., below 0.05) are marked in bold, which in this case applies only to simplicity, ER, graph density, and graph entropy. Notably, the proposed EC methods outperform the rest on these metrics. 

\renewcommand{\thefootnote}{\alph{footnote}}
\setcounter{footnote}{0}
\renewcommand{\footnoterule}{}
\begin{table}[htb!]
\centering
\caption{Average ranking of each clustering method 1-12 (including no clustering), along each metric, over 8 event logs.}
\label{tab:average_rankings}
\begin{minipage}{\textwidth} 
\begin{tabularx}{\textwidth}{l*{8}{>{\centering\arraybackslash}X}}
\toprule
\texttt{method} &
\resizebox{0.09\textwidth}{!}{\texttt{replay\textsubscript{fit}}} &
\resizebox{0.09\textwidth}{!}{\texttt{replay\textsubscript{prec}}} &
\resizebox{0.09\textwidth}{!}{\texttt{align\textsubscript{fit}}\footnotemark[1]} &
\resizebox{0.09\textwidth}{!}{\texttt{align\textsubscript{prec}}\footnotemark[1]} &
\resizebox{0.09\textwidth}{!}{\texttt{simpl\textsubscript{arc}}} &
{\texttt{ER}} &
\resizebox{0.09\textwidth}{!}{\texttt{graph\textsubscript{dens}}} &
\resizebox{0.09\textwidth}{!}{\texttt{graph\textsubscript{entr}}} \\
\midrule
\texttt{full log} & \cellcolor{green!43} 7.25 & \cellcolor{green!47} 6.75 & \cellcolor{green!53} 6.17 & \cellcolor{green!59} 5.50 & \cellcolor{green!4} 11.50 & \cellcolor{green!0} 12.00 & \cellcolor{green!9} 11.00 & \cellcolor{green!0} 12.00 \\ \hline

\texttt{random} & \cellcolor{green!34} 8.17 & \cellcolor{green!33} 8.33 & \cellcolor{green!28} 8.88 & \cellcolor{green!30} 8.62 & \cellcolor{green!27} 9.00 & \cellcolor{green!15} 10.33 & \cellcolor{green!30} 8.67 & \cellcolor{green!21} 9.67 \\

\texttt{frequency\textsubscript{km++}}  & \cellcolor{green!67} \underline{\textbf{4.62}} & \cellcolor{green!73} \underline{\textbf{3.88}} & \cellcolor{green!60} 5.33 & \cellcolor{green!83} \underline{\textbf{2.83}} & \cellcolor{green!51} 6.38 & \cellcolor{green!34} 8.25 & \cellcolor{green!53} 6.12 & \cellcolor{green!67} 4.62 \\

\texttt{trace2vec\textsubscript{km++}} & \cellcolor{green!41} 7.38 & \cellcolor{green!67} 4.62 & \cellcolor{green!39} 7.67 & \cellcolor{green!69} 4.33 & \cellcolor{green!54} 6.00 & \cellcolor{green!26} 9.12 & \cellcolor{green!34} 8.25 & \cellcolor{green!55} 5.88 \\ \hline

\texttt{actitrac\textsubscript{dist}} & \cellcolor{green!37} 7.88 & \cellcolor{green!56} 5.75 & \cellcolor{green!59} 5.50 & \cellcolor{green!54} 6.00 & \cellcolor{green!48} 6.62 & \cellcolor{green!37} 7.88 & \cellcolor{green!30} 8.62 & \cellcolor{green!41} 7.38 \\

\texttt{actitrac\textsubscript{freq}} & \cellcolor{green!39} 7.62 & \cellcolor{green!34} 8.25 & \cellcolor{green!46} 6.83 & \cellcolor{green!42} 7.33 & \cellcolor{green!31} 8.50 & \cellcolor{green!31} 8.50 & \cellcolor{green!32} 8.38 & \cellcolor{green!29} 8.75 \\ \hline

\texttt{EC\textsubscript{++}} & \cellcolor{green!67} \underline{\textbf{4.62}} & \cellcolor{green!38} 7.75 & \cellcolor{green!48} 6.67 & \cellcolor{green!39} 7.67 & \cellcolor{green!59} 5.50 & \cellcolor{green!90} \underline{\textbf{2.00}} & \cellcolor{green!69} 4.38 & \cellcolor{green!69} 4.38 \\

\texttt{EC\textsubscript{++norm}} & \cellcolor{green!52} 6.25 & \cellcolor{green!41} 7.38 & \cellcolor{green!46} 6.83 & \cellcolor{green!34} 8.17 & \cellcolor{green!70} 4.25 & \cellcolor{green!89} 2.12 & \cellcolor{green!78} 3.38 & \cellcolor{green!73} \underline{\textbf{3.88}} \\

\texttt{EC\textsubscript{rand}} & \cellcolor{green!53} 6.12 & \cellcolor{green!47} 6.75 & \cellcolor{green!60} 5.33 & \cellcolor{green!40} 7.50 & \cellcolor{green!76} \underline{\textbf{3.62}} & \cellcolor{green!82} 2.88 & \cellcolor{green!82} \underline{\textbf{2.88}} & \cellcolor{green!68} 4.50 \\

\texttt{EC\textsubscript{split}\textsubscript{++}} & \cellcolor{green!46} 6.88 & \cellcolor{green!52} 6.25 & \cellcolor{green!34} 8.17 & \cellcolor{green!53} 6.17 & \cellcolor{green!61} 5.25 & \cellcolor{green!54} 6.00 & \cellcolor{green!52} 6.25 & \cellcolor{green!45} 7.00 \\

\texttt{EC\textsubscript{split}\textsubscript{++norm}} & \cellcolor{green!52} 6.25 & \cellcolor{green!54} 6.00 & \cellcolor{green!53} 6.17 & \cellcolor{green!45} 7.00 & \cellcolor{green!56} 5.75 & \cellcolor{green!69} 4.38 & \cellcolor{green!57} 5.62 & \cellcolor{green!57} 5.62 \\ 

\texttt{EC\textsubscript{split}\textsubscript{rand}} & \cellcolor{green!65} 4.75 & \cellcolor{green!47} 6.75 & \cellcolor{green!65} \underline{\textbf{4.83}} & \cellcolor{green!42} 7.33 & \cellcolor{green!52} 6.25 & \cellcolor{green!69} 4.38 & \cellcolor{green!67} 4.62 & \cellcolor{green!72} 4.00 \\
\hline \hline

\texttt{Friedman $\chi^2$} & 11.38 & 11.54 & 6.97 & 13.41 & 30.35 & 76.25 & 41.38 & 46.10\\
\texttt{p-value (raw)}  & 0.412 & 0.399 & 0.801 & 0.267 & \textbf{0.001} & \textbf{0.000} & \textbf{0.000} & \textbf{0.000} \\
\texttt{p-value (Holm)}  & 1.000 & 1.000 & 1.000 & 1.000 & \textbf{0.007} & \textbf{0.000} & \textbf{0.000} & \textbf{0.000} \\
\bottomrule
\end{tabularx}
\vspace{-0.3cm}
\footnotetext[1]{\scriptsize{Methods \texttt{align\textsubscript{fit}} and \texttt{align\textsubscript{prec}} did not run on \texttt{BPIC12} and \texttt{BPIC15} sets (out-of-time).}}
\end{minipage}
\end{table}

Based on the ER scores, we can conclude that EC clustering leads to the most clusters, whose DFGs contain the least infrequent DFs , which is expected given that it optimizes for this property. Metrics reflecting the simplicity of DFGs within each cluster, as well as the arc degree of Petri nets discovered using the inductive miner, also rank EC methods highest. The splitting variant of EC generally works worse than the regular EC method, which directly initializes the correct number of clusters, though there are some exceptions. Although statistical significance could not be established across all datasets, it is evident that EC methods tend to perform worse on precision-based metrics—particularly when compared to activity frequency-based clustering.

\begin{figure}[tp] 
    \centering
    \begin{subfigure}{0.75\textwidth}
        \centering
        \includegraphics[width=\linewidth]{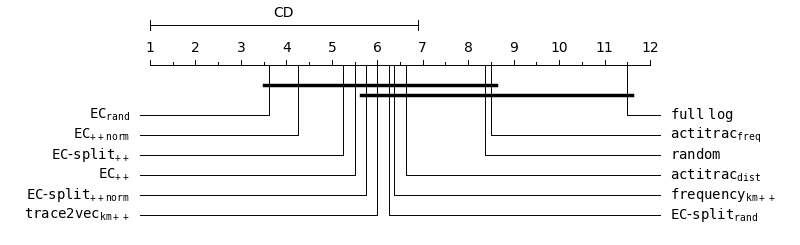}
        \caption{Critical difference diagram for the metric arc degree simplicity.}
        \label{fig:CD_simplicity}
    \end{subfigure}
    
    \vspace{0.1cm} 

    \begin{subfigure}{0.75\textwidth}
        \centering
        \includegraphics[width=\linewidth]{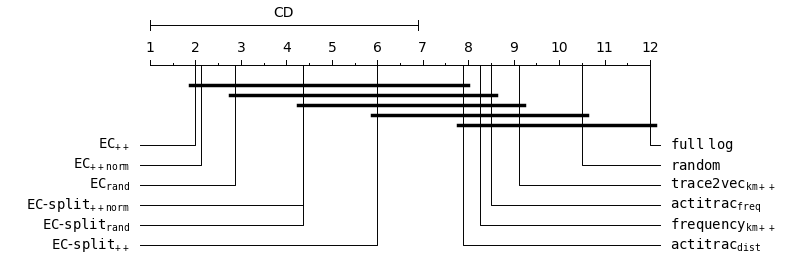}
        \caption{Critical difference diagram for the metric ER.}
        \label{fig:CD_ER}
    \end{subfigure}
    
    \vspace{0.1cm} 
    
    \begin{subfigure}{0.75\textwidth}
        \centering
        \includegraphics[width=\linewidth]{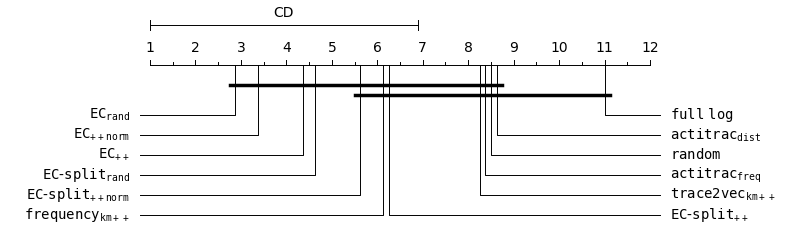}
        \caption{Critical difference diagram for the metric graph density.}
        \label{fig:CD_GD}
    \end{subfigure}
    
    \vspace{0.1cm} 
    
    \begin{subfigure}{0.75\textwidth}
        \centering
        \includegraphics[width=\linewidth]{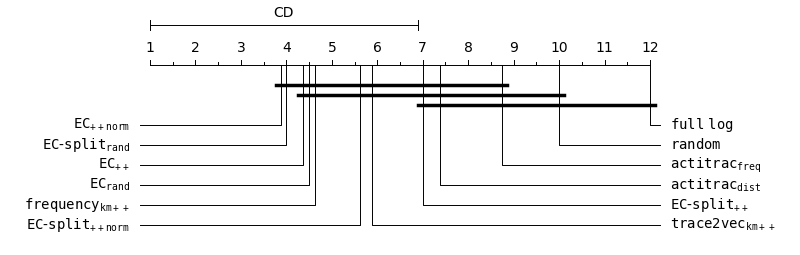}
        \caption{Critical difference diagram for the metric graph entropy.}
        \label{fig:CD_GE}
    \end{subfigure}
    
    \caption{Critical difference diagrams for different metrics.}
    \label{fig:CD_all}
\end{figure}

Figures \ref{fig:CD_simplicity} to \ref{fig:CD_GE} display the critical difference (CD) diagrams for simplicity (of discovered Petri nets), as well as the ER, graph density, and graph entropy of the corresponding DFGs. Notably, many methods are connected in the diagrams, meaning their differences are not statistically significant. In fact, most benchmark clustering methods do not perform significantly better than random clustering or no clustering at all. However, the best-performing EC variants demonstrate a statistically significant advantage in ER and graph entropy compared to these naive approaches. Since the different initialization techniques for EC clustering are all connected in the diagrams, we cannot conclude that any one approach is significantly superior to the others. This could indicate that the methods are less initialization-sensitive than initially thought.

\section{Discussion \& limitations}
\label{sec:discussion}

The experimental results indicate that our entropic clustering approaches lead to more structured clusters when taking a stochastic viewpoint over the log (i.e. whose underlying DFGs contain fewer infrequent DFs), as evidenced by lower ER scores. This, in turn, generally results in simpler downstream process models, both in terms of directly-follows graphs (DFGs) and Petri nets discovered using the Inductive Miner. It is, however, impossible to draw general conclusions about the quality of clustering methods concerning the fitness of the discovered Petri nets. In general, clustering based on activity frequency and trace2vec tends to produce more precise models, than our EC-based methods. These lower precision scores on the discovered Petri nets could perhaps be attributed to their optimization for DFGs, only focusing on DFs and therefore potentially generalizing too much behavior, potentially overlooking aspects such as parallelism. Since conformance checking in our experiments was performed on models mined using the Inductive Miner, the power and limitations of this discovery algorithm also influence the results.

Based on these findings, we can offer practical guidance on when to apply Entropic Clustering versus traditional approaches. If the primary analytical goal is to produce precise, replayable Petri nets where structural constraints (such as parallelism and exact traces) are paramount, traditional frequency-based or vector-based clustering methods may be more suitable. However, Entropic Clustering is the superior choice when the analysis relies on Directly-Follows Graphs, as is common in commercial process mining tools, and when the user requires an understanding of the dominant probabilistic behaviors. By penalizing low-probability transitions, EC is particularly effective at untangling ``spaghetti'' processes into coherent, stochastically valid subgroups, making it ideal for exploratory analysis and interpretability-focused tasks.

In our current experimental setup, we did not include other stochastic conformance checking metrics. Other conformance checking metrics, such as using the Earth mover's distance~\cite{Leemans2019}, would be most meaningful if we were not using fully fitting (non-filtered) DFGs as the discovered model. We would therefore have to decide on one stochastic discovery technique, which could have a large influence on the end results. Or, to properly account for this, try different techniques (and filter) leading to exponentially more experiments. We opted for ER as our primary measure because it captures the (stochastic) the lack of structure in clusters in a general way, independent of the possible downstream discovery.

Regarding the computation of this metric, we recognize that a lower numerical cut-off (allowing smaller probabilities) would produce more precise absolute ER scores for some logs. However, our experiments are designed to compare the relative performance of clustering methods. Since this probability floor $\varepsilon$ is applied uniformly across all methods, it acts as a consistent penalty for extreme outliers and is unlikely to change the relative ranking of the algorithms or the overall conclusions.

While statistical ranking tests provide a more robust evaluation of different methods compared to simply reporting scores per event log, they do not indicate which method works best for specific types of processes. Another limitation of ranking-based evaluation is that if all methods perform poorly for a particular log, the ranking becomes somewhat arbitrary.

Unlike methods such as k-means, our approach does not rely on shifting cluster centroids to reach an equilibrium, as defining efficient cluster centers or medoids for process behavior is not straightforward. Consequently, the EC method could still remain somewhat sensitive to initialization in its current form. Our \texttt{++} and \texttt{++norm} initializations still select the first seed randomly, which can lead to suboptimal starting conditions, and for the recursive \texttt{EC}\textsubscript{split} variant at each split. A more computationally expensive approach, incorporating all pairwise distances, could be explored to mitigate this issue. Additionally, the \texttt{++} and \texttt{++norm} initializations, despite the latter’s normalization to account for loops, tend to favor outlier traces as initial seeds. While this may be beneficial in some cases, it could be suboptimal for certain applications. The lack of statistically significant differences between initialization methods may point to a lower-than-expected sensitivity to initialization. Nevertheless, it might be interesting to experiment with a k-medoids-style approach, where the initialisation is updated over multiple runs. This could be done by, e.g., looking at the trace with the highest play-out probability (or lowest ER) with the current cluster's DFG, in each cluster. 

\section{Conclusion \& future work}
\label{sec:conclusion}

In this work, we introduced a trace clustering paradigm focused on reducing the complexity of underlying clusters from a stochastic process model perspective. Our entropic clustering techniques actively utilize the directly-follows graph (DFG) within each cluster, assigning traces by minimizing entropic relevance. Through extensive experiments on real-life event logs, we demonstrated that our methods outperform alternatives in minimizing entropic relevance while also producing less complex downstream process models, both in terms of DFGs and discovered Petri nets. Additionally, our experiments highlight how the perception of clustering quality changes depending on whether stochasticity is taken into account. Our approach is highly efficient, optimizing for commonly used DFGs, making it a viable candidate for integration into commercial process mining tools.

For future work, a larger-scale experiment could be conducted, potentially using controlled synthetic data. This would allow us to systematically analyze which types of process behaviors and challenges affect different clustering techniques. A simple but useful extension could involve introducing a robust version of the algorithm that detects and handles outliers, as they likely skew clustering results in the current approach. Further extensions could expand the model-driven stochastic clustering paradigm by incorporating other process model types, discovery algorithms, and stochastic conformance checking techniques. We opted for the current combination due to its efficiency and widespread adoption in commercial tools. Our algorithm (see Alg.~\ref{alg:EC}) updates each cluster’s DFG incrementally when assigning traces. An alternative approach could forgo this update and use the DFGs as they are, relying on an ER metric implementation~\cite{Alkhammash2022} including an appropriate background model for non-fitting traces. Another avenue for improvement lies in the initialization phase. Seed selection could be guided by expert knowledge, if available. Given the scalability of our method, running the clustering multiple times and selecting the best result could help mitigate volatility due to initialization. This could also be done in a k-medoids-style approach, by running the clustering algorithm multiple times (with updated initialization) until no change is detected. The new initial clusters could in each iteration be selected as the trace with the lowest ER score (with the cluster's DFG). Additionally, since determining the correct number of clusters upfront is often impractical, a variant that dynamically adjusts the number of clusters, based on a maximum ER threshold, for example, could be explored. Finally, our approach could be extended to incorporate additional process features, such as the resource performing an activity or sensor measurements. One way to achieve this would be to adjust probability calculations based on these features. For categorical attributes, probabilities could be derived from frequency distributions in the current traces, while for continuous attributes, they could be estimated using a fitted (normal) distribution.

\section*{Acknowledgments}
This work was supported in part by the Research Foundation Flanders (FWO) under Project 1294325N as well as grant number G039923N, and Internal Funds KU Leuven under grant number C14/23/031.

\bibliographystyle{elsarticle-num}
\bibliography{cas-refs}

\newpage
\appendix
\section{Results for the Individual Event Logs}
\label{sec:FullResults}

In this appendix, we present the results for each event log separately. Columns correspond to the different metrics and rows with the different clustering techniques. We report weighted averages, i.e., to calculate the metric value for a clustering, the measurements for each cluster are weighted by the number of cases in that cluster. Except for the column displaying \texttt{ER\textsubscript{sum}} which displays the total sum of the ER over all clusters (and all traces within those clusters). The weighted average \texttt{ER} is the same as the values in the \texttt{ER\textsubscript{sum}} column divided by the total number of cases in the event log. In addition, we provide figures for each cluster’s DFG in our online repository, together with the script used to generate them, which allows readers to visually compare the resulting models across clusters and methods.

\begin{table}[htbp]
    \centering
    \caption{Weighted averages \texttt{Helpdesk}}
    \resizebox{\linewidth}{!}{\begin{tabular}{lccccccccc} 
\toprule
method & \texttt{replay\textsubscript{fit}} & \texttt{replay\textsubscript{prec}} & \texttt{align\textsubscript{fit}} & \texttt{align\textsubscript{prec}} & \texttt{simpl\textsubscript{arc}} & \texttt{ER\textsubscript{av}} & \texttt{ER\textsubscript{sum}} & \texttt{graph\textsubscript{dens}} & \texttt{graph\textsubscript{entr}} \\
\midrule
\texttt{full log} &  0.989 & 0.781 & 0.978 & 0.781 & 0.682 & 3.618 & 16570 & 0.279 & 12.729 \\
\hline
\texttt{random} &  0.992 & 0.695 & 0.984 & 0.695 & 0.690 & 2.754 & 12614 & 0.268 & 7.912 \\
\texttt{frequency\textsubscript{km++}} &  \underline{\textbf{0.995}} & 0.831 & \underline{\textbf{0.990}} & 0.831 & 0.686 & 2.670 & 12230 & 0.259 & \underline{\textbf{4.779}} \\
\texttt{trace2vec\textsubscript{km++}} &  0.983 & 0.845 & 0.977 & 0.845 & 0.696 & 2.744 & 12566 & 0.243 & 6.367 \\
\hline
\texttt{actitrac\textsubscript{dist}} &  0.993 & 0.659 & 0.986 & 0.659 & 0.682 & 2.434 & 11150 & 0.253 & 7.743 \\
\texttt{actitrac\textsubscript{freq}} &  0.919 & \underline{\textbf{0.885}} & 0.918 & \underline{\textbf{0.886}} & 0.686 & 3.026 & 13858 & 0.266 & 8.807 \\ \hline
\texttt{EC\textsubscript{++}} &  0.974 & 0.607 & 0.973 & 0.607 & \underline{\textbf{0.717}} & 2.108 & 9656 & \underline{\textbf{0.194}} & 5.846 \\
\texttt{EC\textsubscript{++norm}} &  0.974 & 0.596 & 0.973 & 0.596 & 0.716 & \underline{\textbf{2.105}} & \underline{\textbf{9641}} & 0.198 & 5.839 \\
\texttt{EC\textsubscript{rand}} &  0.969 & 0.664 & 0.968 & 0.664 & 0.699 & 2.118 & 9702 & 0.216 & 6.094 \\
\texttt{EC\textsubscript{split}\textsubscript{++}} &  0.967 & 0.750 & 0.954 & 0.750 & 0.702 & 2.379 & 10894 & 0.276 & 7.241 \\
\texttt{EC\textsubscript{split}\textsubscript{++norm}} &  0.971 & 0.757 & 0.961 & 0.757 & 0.698 & 2.387 & 10930 & 0.292 & 7.152 \\
\texttt{EC\textsubscript{split}\textsubscript{rand}} &  0.987 & 0.757 & 0.977 & 0.757 & 0.692 & 2.482 & 11369 & 0.292 & 7.037 \\
\bottomrule
\end{tabular}}
\end{table}

\begin{table}[htbp]
    \centering
    \caption{Weighted averages \texttt{RTFM}}
    \resizebox{\linewidth}{!}{\begin{tabular}{lccccccccc} 
\toprule
method & \texttt{replay\textsubscript{fit}} & \texttt{replay\textsubscript{prec}} & \texttt{align\textsubscript{fit}} & \texttt{align\textsubscript{prec}} & \texttt{simpl\textsubscript{arc}} & \texttt{ER\textsubscript{av}} & \texttt{ER\textsubscript{sum}} &\texttt{graph\textsubscript{dens}} & \texttt{graph\textsubscript{entr}} \\
\midrule
\texttt{full log} &  0.929 & 0.813 & 0.878 & 0.813 & 0.630 & 2.935 & 441317 & 0.500 & 11.428 \\
\hline
\texttt{random} &  0.978 & 0.660 & 0.971 & 0.660 & 0.676 & 1.453 & 218421 & 0.419 & 10.687 \\
\texttt{frequency\textsubscript{km++}} &  \underline{\textbf{0.997}} & 0.880 & \underline{\textbf{0.993}} & 0.880 & 0.693 & 1.708 & 256904 & \underline{\textbf{0.283}} & \underline{\textbf{7.407}} \\
\texttt{trace2vec\textsubscript{km++}} &  0.991 & \underline{\textbf{0.925}} & 0.987 & \underline{\textbf{0.925}} & 0.746 & 1.229 & 184810 & 0.341 & 9.038 \\
\hline
\texttt{actitrac\textsubscript{dist}} &  0.994 & 0.825 & 0.992 & 0.825 & 0.768 & 0.961 & 144564 & 0.369 & 5.157 \\
\texttt{actitrac\textsubscript{freq}} &  0.968 & 0.752 & 0.955 & 0.752 & 0.681 & 1.250 & 187920 & 0.403 & 8.415 \\ \hline
\texttt{EC\textsubscript{++}} &  0.995 & 0.818 & 0.988 & 0.818 & \underline{\textbf{0.776}} & \underline{\textbf{0.730}} & 109777 & 0.333 & 8.567 \\
\texttt{EC\textsubscript{++norm}} &  0.946 & 0.682 & 0.950 & 0.682 & 0.726 & \underline{\textbf{0.730}} & \underline{\textbf{109721}} & 0.300 & 7.720 \\
\texttt{EC\textsubscript{rand}} &  0.995 & 0.792 & 0.988 & 0.792 & 0.774 & 0.731 & 109866 & 0.326 & 8.425 \\
\texttt{EC\textsubscript{split}\textsubscript{++}} &  0.981 & 0.735 & 0.980 & 0.735 & 0.742 & 0.960 & 144290 & 0.388 & 7.756 \\
\texttt{EC\textsubscript{split}\textsubscript{++norm}} &  0.981 & 0.687 & 0.980 & 0.687 & 0.676 & 0.960 & 144333 & 0.366 & 7.780 \\
\texttt{EC\textsubscript{split}\textsubscript{rand}} &  0.981 & 0.665 & 0.980 & 0.665 & 0.707 & 0.958 & 144564 & 0.355 & 7.733 \\

\bottomrule
\end{tabular}}
\end{table}

\begin{table}[htbp]
    \centering
    \caption{Weighted averages \texttt{Hospital billing}}
    \resizebox{\linewidth}{!}{\begin{tabular}{lccccccccc} 
\toprule
method & \texttt{replay\textsubscript{fit}} & \texttt{replay\textsubscript{prec}} & \texttt{align\textsubscript{fit}} & \texttt{align\textsubscript{prec}} & \texttt{simpl\textsubscript{arc}} & \texttt{ER\textsubscript{av}} & \texttt{ER\textsubscript{sum}} & \texttt{graph\textsubscript{dens}} & \texttt{graph\textsubscript{entr}} \\
\midrule
\texttt{full log} &  0.990 & 0.591 & 0.975 & 0.591 & 0.629 & 3.450 & 345039 & 0.416 & 18.223 \\
\hline
\texttt{random} &  0.993 & 0.463 & 0.979 & 0.463 & 0.639 & 2.307 & 230692 & 0.318 & 15.437 \\
\texttt{frequency\textsubscript{km++}} &  0.995 & 0.637 & 0.987 & 0.637 & 0.648 & 2.156 & 215589 & 0.258 & 11.383 \\
\texttt{trace2vec\textsubscript{km++}} &  0.984 & 0.472 & 0.964 & 0.472 & 0.635 & 2.410 & 241045 & 0.275 & \underline{\textbf{10.290}} \\
\hline
\texttt{actitrac\textsubscript{dist}} &  0.988 & \underline{\textbf{0.787}} & 0.957 & \underline{\textbf{0.787}} & \underline{\textbf{0.685}} & 1.942 & 194174 & 0.337 & 11.546 \\
\texttt{actitrac\textsubscript{freq}} &  0.991 & 0.627 & 0.988 & 0.627 & 0.631 & 1.968 & 196769 & 0.304 & 12.834 \\ \hline
\texttt{EC\textsubscript{++}} &  \underline{\textbf{0.998}} & 0.615 & \underline{\textbf{0.995}} & 0.618 & 0.658 & 1.237 & 123719 & 0.249 & 13.752 \\
\texttt{EC\textsubscript{++norm}} &  0.995 & 0.691 & 0.985 & 0.697 & 0.670 & \underline{\textbf{1.232}} & \underline{\textbf{123175}} & 0.240 & 13.214 \\
\texttt{EC\textsubscript{rand}} &  0.995 & 0.674 & 0.984 & 0.677 & 0.671 & 1.239 & 123922 & 0.239 & 13.512 \\
\texttt{EC\textsubscript{split}\textsubscript{++}} &  0.988 & 0.669 & 0.967 & 0.669 & 0.666 & 1.356 & 135583 & 0.309 & 14.327 \\
\texttt{EC\textsubscript{split}\textsubscript{++norm}} &  0.983 & 0.642 & 0.963 & 0.642 & 0.663 & 1.346 & 134598 & 0.237 & 11.695 \\
\texttt{EC\textsubscript{split}\textsubscript{rand}} &  0.984 & 0.701 & 0.963 & 0.702 & 0.657 & 1.355 & 135463 & \underline{\textbf{0.230}} & 11.167 \\
\bottomrule
\end{tabular}}
\end{table}

\begin{table}[htbp]
    \centering
    \caption{Weighted averages \texttt{Sepsis}}
    \resizebox{\linewidth}{!}{\begin{tabular}{lccccccccc} 
\toprule
method & \texttt{replay\textsubscript{fit}} & \texttt{replay\textsubscript{prec}} & \texttt{align\textsubscript{fit}} & \texttt{align\textsubscript{prec}} & \texttt{simpl\textsubscript{arc}} & \texttt{ER\textsubscript{av}} & \texttt{ER\textsubscript{sum}} & \texttt{graph\textsubscript{dens}} & \texttt{graph\textsubscript{entr}} \\
\midrule
\texttt{full log} &  0.967 & 0.619 & 0.962 & 0.619 & 0.593 & 24.444 & 25666 & 0.441 & 23.580 \\
\hline

\texttt{random} &  0.954 & 0.663 & 0.949 & \underline{\textbf{0.679}} & 0.601 & 23.983 & 25182 & 0.364 & 21.753 \\
\texttt{frequency\textsubscript{km++}} &  0.978 & \underline{\textbf{0.672}} & 0.962 & 0.674 & 0.614 & 22.594 & 23724 & 0.342 & 18.925 \\
\texttt{trace2vec\textsubscript{km++}} &  0.972 & 0.602 & 0.956 & 0.602 & \underline{\textbf{0.629}} & 23.175 & 24334 & 0.499 & \underline{\textbf{18.212}} \\
\hline
\texttt{actitrac\textsubscript{dist}} &  0.968 & 0.567 & 0.967 & 0.575 & 0.606 & 22.695 & 23829 & 0.367 & 20.646 \\
\texttt{actitrac\textsubscript{freq}} &  0.976 & 0.610 & 0.969 & 0.610 & 0.623 & 22.621 & 23752 & 0.379 & 20.710 \\ \hline
\texttt{EC\textsubscript{++}} &  0.961 & 0.532 & 0.948 & 0.536 & 0.601 & \underline{\textbf{21.163}} & \underline{\textbf{22221}} & 0.326 & 18.841 \\
\texttt{EC\textsubscript{++norm}} &  0.974 & 0.539 & 0.958 & 0.543 & 0.603 & 21.324 & 22391 & 0.327 & 18.977 \\
\texttt{EC\textsubscript{rand}} &  0.979 & 0.554 & 0.960 & 0.558 & 0.610 & 21.220 & 22281 & \underline{\textbf{0.317}} & 18.697 \\
\texttt{EC\textsubscript{split}\textsubscript{++}} &  0.964 & 0.506 & 0.953 & 0.512 & 0.608 & 22.243 & 23355 & 0.338 & 19.563 \\
\texttt{EC\textsubscript{split}\textsubscript{++norm}} &  0.971 & 0.486 & 0.959 & 0.490 & 0.596 & 22.039 & 23141 & 0.325 & 19.497 \\
\texttt{EC\textsubscript{split}\textsubscript{rand}} &  \underline{\textbf{0.986}} & 0.455 & \underline{\textbf{0.971}} & 0.457 & 0.612 & 22.040 & 23142 & 0.332 & 19.332 \\
\bottomrule
\end{tabular}}
\end{table}

\begin{table}[htbp]
    \centering
    \caption{Weighted averages \texttt{BPIC13 incidents}}
    \resizebox{\linewidth}{!}{\begin{tabular}{lccccccccc} 
\toprule
method & \texttt{replay\textsubscript{fit}} & \texttt{replay\textsubscript{prec}} & \texttt{align\textsubscript{fit}} & \texttt{align\textsubscript{prec}} & \texttt{simpl\textsubscript{arc}} & \texttt{ER\textsubscript{av}} & \texttt{ER\textsubscript{sum}} & \texttt{graph\textsubscript{dens}} & \texttt{graph\textsubscript{entr}} \\
\midrule
\texttt{full log} &  0.966 & 0.841 & 0.957 & 0.841 & 0.733 & 10.728 & 81036 & 0.533 & 3.668 \\
\hline
\texttt{random} &  0.959 & 0.866 & 0.953 & 0.866 & 0.734 & 10.498 & 79305 & 0.586 & 3.558 \\
\texttt{frequency\textsubscript{km++}} &  0.987 & \underline{\textbf{0.947}} & 0.930 & \underline{\textbf{0.960}} & 0.770 & 9.704 & 73301 & 0.522 & 3.079 \\
\texttt{trace2vec\textsubscript{km++}} &  0.945 & 0.877 & 0.954 & 0.877 & 0.697 & 10.452 & 78958 & 0.653 & 3.657 \\
\hline
\texttt{actitrac\textsubscript{dist}} &  0.988 & 0.844 & 0.983 & 0.844 & 0.722 & 9.865 & 74520 & 0.528 & 3.287 \\
\texttt{actitrac\textsubscript{freq}} &  0.988 & 0.775 & 0.983 & 0.775 & 0.722 & 9.867 & 74536 & 0.516 & 3.305 \\ \hline
\texttt{EC\textsubscript{++}} &  0.995 & 0.839 & 0.950 & 0.854 & 0.766 & \underline{\textbf{9.191}} & 69431 & 0.540 & 2.879 \\
\texttt{EC\textsubscript{++norm}} &  0.995 & 0.866 & 0.986 & 0.866 & \underline{\textbf{0.784}} & \underline{\textbf{9.191}} & \underline{\textbf{69429}} & \underline{\textbf{0.455}} & \underline{\textbf{2.544}} \\
\texttt{EC\textsubscript{rand}} &  0.991 & 0.833 & 0.979 & 0.833 & 0.771 & 9.445 & 71346 & 0.513 & 2.953 \\
\texttt{EC\textsubscript{split}\textsubscript{++}} &  \underline{\textbf{0.997}} & 0.909 & 0.964 & 0.915 & 0.769 & 9.767 & 73778 & 0.537 & 3.319 \\
\texttt{EC\textsubscript{split}\textsubscript{++norm}} &  \underline{\textbf{0.997}} & 0.926 & 0.991 & 0.926 & 0.772 & 9.506 & 71809 & 0.561 & 3.143 \\
\texttt{EC\textsubscript{split}\textsubscript{rand}} &  \underline{\textbf{0.997}} & 0.849 & \underline{\textbf{0.994}} & 0.849 & 0.748 & 9.380 & 70854 & 0.504 & 3.037 \\
\bottomrule
\end{tabular}}
\end{table}

\begin{table}[htbp]
    \centering
    \caption{Weighted averages \texttt{BPIC13 closed problems}}
    \resizebox{\linewidth}{!}{\begin{tabular}{lccccccccc} 
\toprule
method & \texttt{replay\textsubscript{fit}} & \texttt{replay\textsubscript{prec}} & \texttt{align\textsubscript{fit}} & \texttt{align\textsubscript{prec}} & \texttt{simpl\textsubscript{arc}} & \texttt{ER\textsubscript{av}} & \texttt{ER\textsubscript{sum}} & \texttt{graph\textsubscript{dens}} & \texttt{graph\textsubscript{entr}} \\
\midrule
\texttt{full log} &  0.997 & \underline{\textbf{0.971}} & 0.990 & \underline{\textbf{0.971}} & 0.750 & 5.047 & 7505 & 0.500 & 3.916 \\
\hline
\texttt{random} &  0.989 & 0.802 & 0.957 & 0.802 & 0.766 & 4.675 & 6952 & 0.380 & 2.455 \\
\texttt{frequency\textsubscript{km++}} &  0.986 & 0.958 & 0.969 & 0.959 & \underline{\textbf{0.783}} & 4.075 & 6060 & 0.452 & 2.346 \\
\texttt{trace2vec\textsubscript{km++}} &  0.993 & 0.941 & 0.977 & 0.941 & 0.776 & 4.717 & 7014 & 0.448 & 2.711 \\
\hline
\texttt{actitrac\textsubscript{dist}} &  0.987 & 0.901 & 0.970 & 0.901 & 0.775 & 4.342 & 6457 & 0.453 & 2.819 \\
\texttt{actitrac\textsubscript{freq}} &  0.986 & 0.796 & 0.966 & 0.796 & 0.770 & 4.567 & 6791 & 0.458 & 3.154 \\ \hline
\texttt{EC\textsubscript{++}} &  0.997 & 0.907 & 0.989 & 0.907 & 0.772 & 3.826 & 5690 & 0.371 & \underline{\textbf{1.622}} \\
\texttt{EC\textsubscript{++norm}} &  0.996 & 0.813 & 0.989 & 0.813 & 0.774 & \underline{\textbf{3.792}} & \underline{\textbf{5639}} & \underline{\textbf{0.337}} & 1.625 \\
\texttt{EC\textsubscript{rand}} &  0.996 & 0.850 & 0.990 & 0.850 & 0.773 & 3.798 & 5648 & 0.344 & 1.953 \\
\texttt{EC\textsubscript{split}\textsubscript{++}} &  0.996 & 0.938 & 0.989 & 0.938 & 0.758 & 4.022 & 5981 & 0.394 & 2.751 \\
\texttt{EC\textsubscript{split}\textsubscript{++norm}} &  \underline{\textbf{0.998}} & 0.813 & \underline{\textbf{0.994}} & 0.813 & 0.768 & 3.921 & 5830 & 0.407 & 2.782 \\
\texttt{EC\textsubscript{split}\textsubscript{rand}} &  0.996 & 0.922 & 0.990 & 0.922 & 0.764 & 3.975 & 5911 & 0.399 & 2.588 \\
\bottomrule
\end{tabular}}
\end{table}

\begin{table}[htbp]
    \centering
    \caption{Weighted averages \texttt{BPIC12}}
    \resizebox{\linewidth}{!}{\begin{tabular}{lccccccccc} 
\toprule
method & \texttt{replay\textsubscript{fit}} & \texttt{replay\textsubscript{prec}} & \texttt{align\textsubscript{fit}} & \texttt{align\textsubscript{prec}} & \texttt{simpl\textsubscript{arc}} & \texttt{ER\textsubscript{av}} & \texttt{ER\textsubscript{sum}} & \texttt{graph\textsubscript{dens}} & \texttt{graph\textsubscript{entr}} \\
\midrule
\texttt{full log} &  \underline{\textbf{0.999}} & 0.150 & / & / & 0.600 & 19.599 & 256496 & 0.211 & 34.259 \\
\hline
\texttt{random} &  0.971 & 0.342 & / & / & 0.625 & 18.803 & 246081 & 0.204 & 32.639 \\
\texttt{frequency\textsubscript{km++}} &  0.973 & 0.532 & / & / & 0.665 & 18.626 & 243757 & 0.189 & 30.663 \\
\texttt{trace2vec\textsubscript{km++}} &  0.971 & 0.535 & / & / & 0.686 & 18.579 & 243138 & 0.172 & 22.166 \\
\hline
\texttt{actitrac\textsubscript{dist}} &  0.944 & 0.397 & / & / & 0.652 & 17.326 & 226741 & 0.191 & 27.323 \\
\texttt{actitrac\textsubscript{freq}} &  0.997 & 0.294 & / & / & 0.643 & 16.963 & 221992 & 0.176 & 26.933 \\ \hline
\texttt{EC\textsubscript{++}} &  0.993 & 0.386 & / & / & 0.640 & 16.798 & 219834 & 0.171 & 24.821 \\
\texttt{EC\textsubscript{++norm}} &  0.993 & 0.369 & / & / & 0.658 & 16.861 & 220660 & 0.163 & 24.149 \\
\texttt{EC\textsubscript{rand}} &  0.964 & 0.680 & / & / & 0.690 & \underline{\textbf{16.841}} & \underline{\textbf{220403}} & \underline{\textbf{0.140}} & 19.965 \\
\texttt{EC\textsubscript{split}\textsubscript{++}} &  0.966 & \underline{\textbf{0.690}} & / & / & \underline{\textbf{0.714}} & 17.030 & 222875 & 0.151 & 20.643 \\
\texttt{EC\textsubscript{split}\textsubscript{++norm}} &  0.965 & 0.622 & / & / & 0.699 & 16.980 & 222212 & 0.151 & 20.009 \\
\texttt{EC\textsubscript{split}\textsubscript{rand}} &  0.969 & 0.624 & / & / & 0.707 & 16.912 & 221330 & 0.151 & \underline{\textbf{18.398}} \\
\bottomrule
\end{tabular}}
\end{table}

\begin{table}[htbp]
    \centering
    \caption{Weighted averages \texttt{BPIC15}}
    \resizebox{\linewidth}{!}{\begin{tabular}{lccccccccc} 
\toprule
method & \texttt{replay\textsubscript{fit}} & \texttt{replay\textsubscript{prec}} & \texttt{align\textsubscript{fit}} & \texttt{align\textsubscript{prec}} & \texttt{simpl\textsubscript{arc}} & \texttt{ER\textsubscript{av}} & \texttt{ER\textsubscript{sum}} & \texttt{graph\textsubscript{dens}} & \texttt{graph\textsubscript{entr}} \\
\midrule
\texttt{full log} &  0.884 & 0.366 & / & / & 0.572 & 26.273 & 148365 & 0.308 & 29.932 \\
\hline
\texttt{random} &  0.901 & 0.415 & / & / & 0.625 & 26.123 & 147466 & 0.235 & 26.660 \\
\texttt{frequency\textsubscript{km++}} &  0.938 & 0.382 & / & / & 0.600 & 24.584 & 138775 & 0.235 & 24.045 \\
\texttt{trace2vec\textsubscript{km++}} &  0.938 & 0.397 & / & / & 0.612 & 24.351 & 137463 & 0.246 & 22.264 \\
\hline
\texttt{actitrac\textsubscript{dist}} &  0.905 & 0.401 & / & / & 0.612 & 25.223 & 142486 & 0.245 & 25.279 \\
\texttt{actitrac\textsubscript{freq}} &  0.930 & 0.364 & / & / & 0.606 & 25.146 & 142026 & 0.244 & 24.257 \\ \hline
\texttt{EC\textsubscript{++}} &  0.937 & 0.397 & / & / & 0.620 & 24.250 & 136892 & 0.223 & \underline{\textbf{20.707}} \\
\texttt{EC\textsubscript{++norm}} &  0.916 & \underline{\textbf{0.469}} & / & / & 0.625 & 24.387 & 137666 & 0.240 & 22.087 \\
\texttt{EC\textsubscript{rand}} &  0.930 & 0.395 & / & / & 0.623 & 24.330 & 137342 & 0.239 & 22.198 \\
\texttt{EC\textsubscript{split}\textsubscript{++}} &  \underline{\textbf{0.943}} & 0.354 & / & / & 0.614 & 24.393 & 137698 & 0.218 & 21.741 \\
\texttt{EC\textsubscript{split}\textsubscript{++norm}} &  0.940 & 0.432 & / & / & \underline{\textbf{0.642}} & \underline{\textbf{24.191}} & \underline{\textbf{136560}} & 0.223 & 20.883 \\
\texttt{EC\textsubscript{split}\textsubscript{rand}} &  0.942 & 0.384 & / & / & 0.616 & 24.249 & 136884 & \underline{\textbf{0.217}} & 20.916 \\
\bottomrule
\end{tabular}}
\end{table}

\clearpage

\section{Cluster count selection (Elbow Analysis)}
\label{sec:Elbow}

\begin{figure}[htbp]
    \centering
    \begin{subfigure}[b]{0.49\textwidth}
        \centering
        \includegraphics[trim={1.5cm 0.6cm 2.5cm 1.2cm}, clip, width=\linewidth]{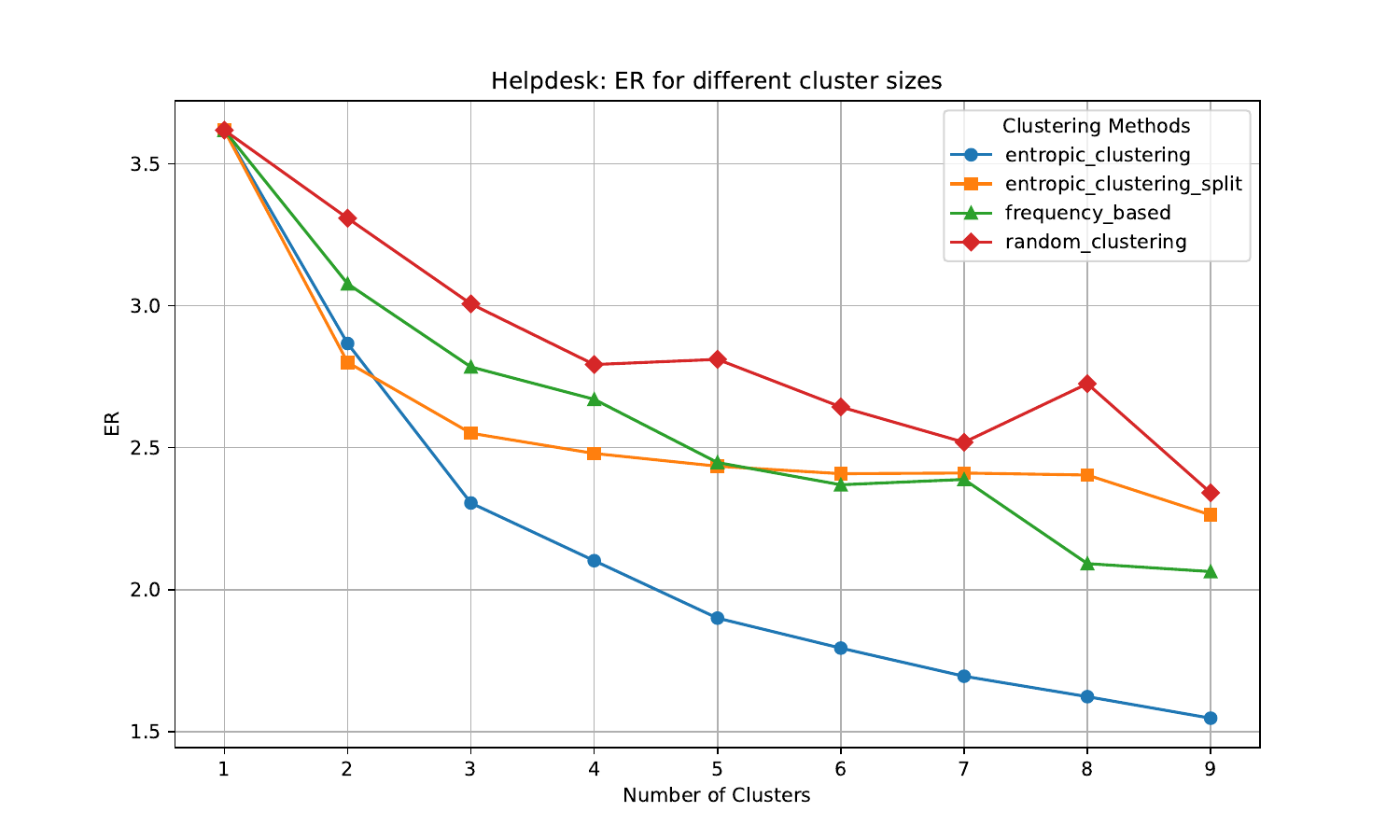}
    \end{subfigure}
    \hfill
    \begin{subfigure}[b]{0.49\textwidth}
        \centering
        \includegraphics[trim={1.5cm 0.6cm 2.5cm 1.2cm}, clip, width=\linewidth]{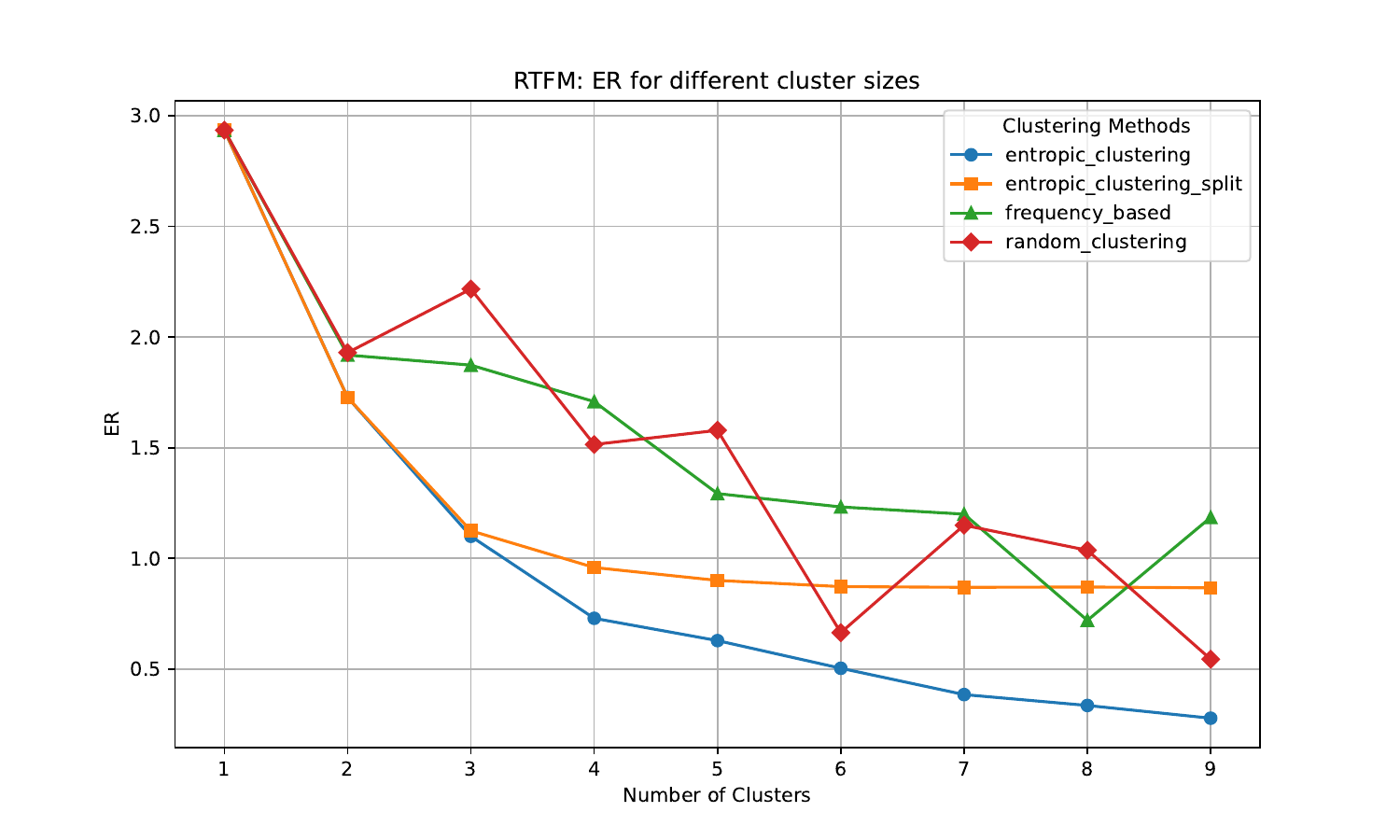}
    \end{subfigure}
    
    \vspace{0.5cm} 
    
    \begin{subfigure}[b]{0.49\textwidth}
        \centering
        \includegraphics[trim={1.5cm 0.6cm 2.5cm 1.2cm}, clip, width=\linewidth]{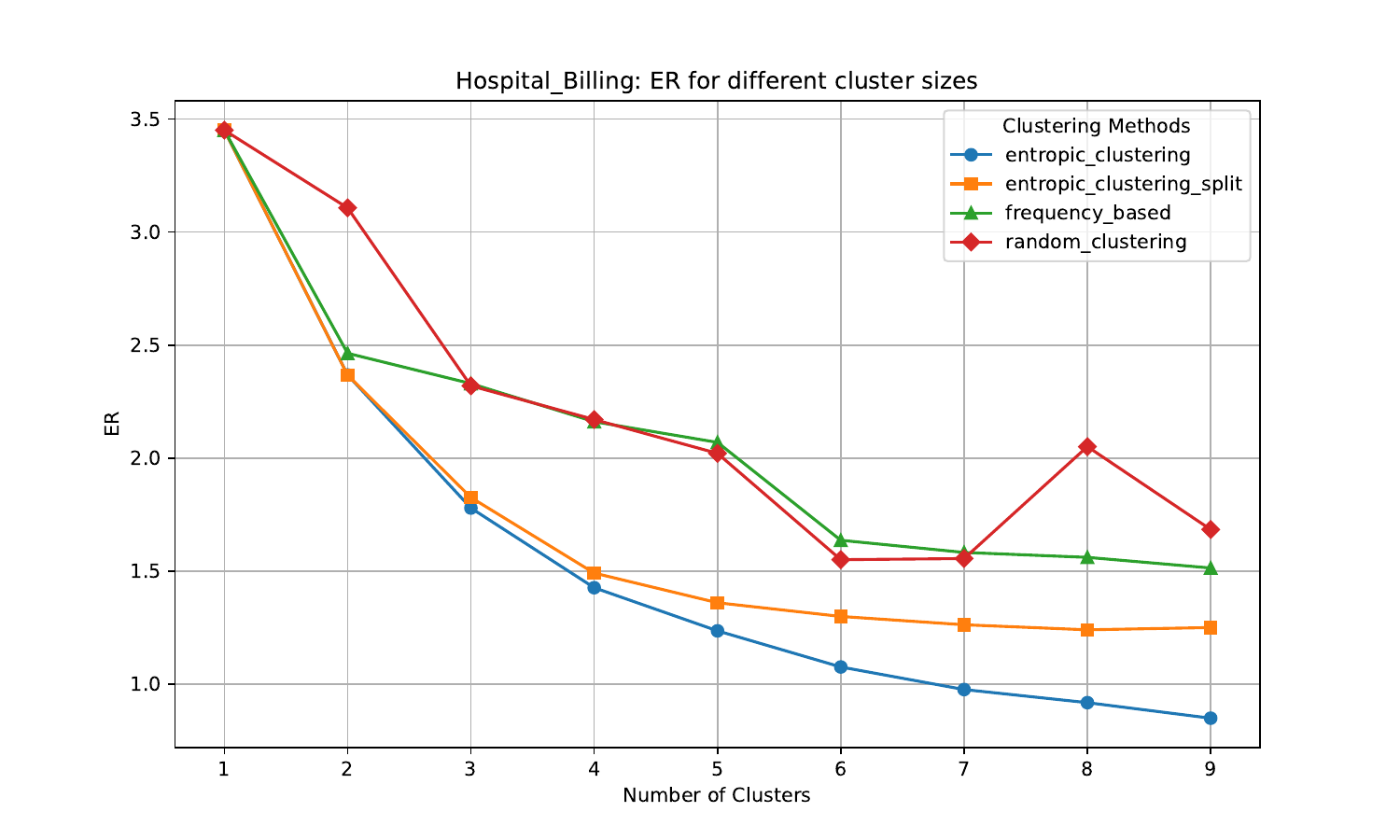}
    \end{subfigure}
    \hfill
    \begin{subfigure}[b]{0.49\textwidth}
        \centering
        \includegraphics[trim={1.5cm 0.6cm 2.5cm 1.2cm}, clip, width=\linewidth]{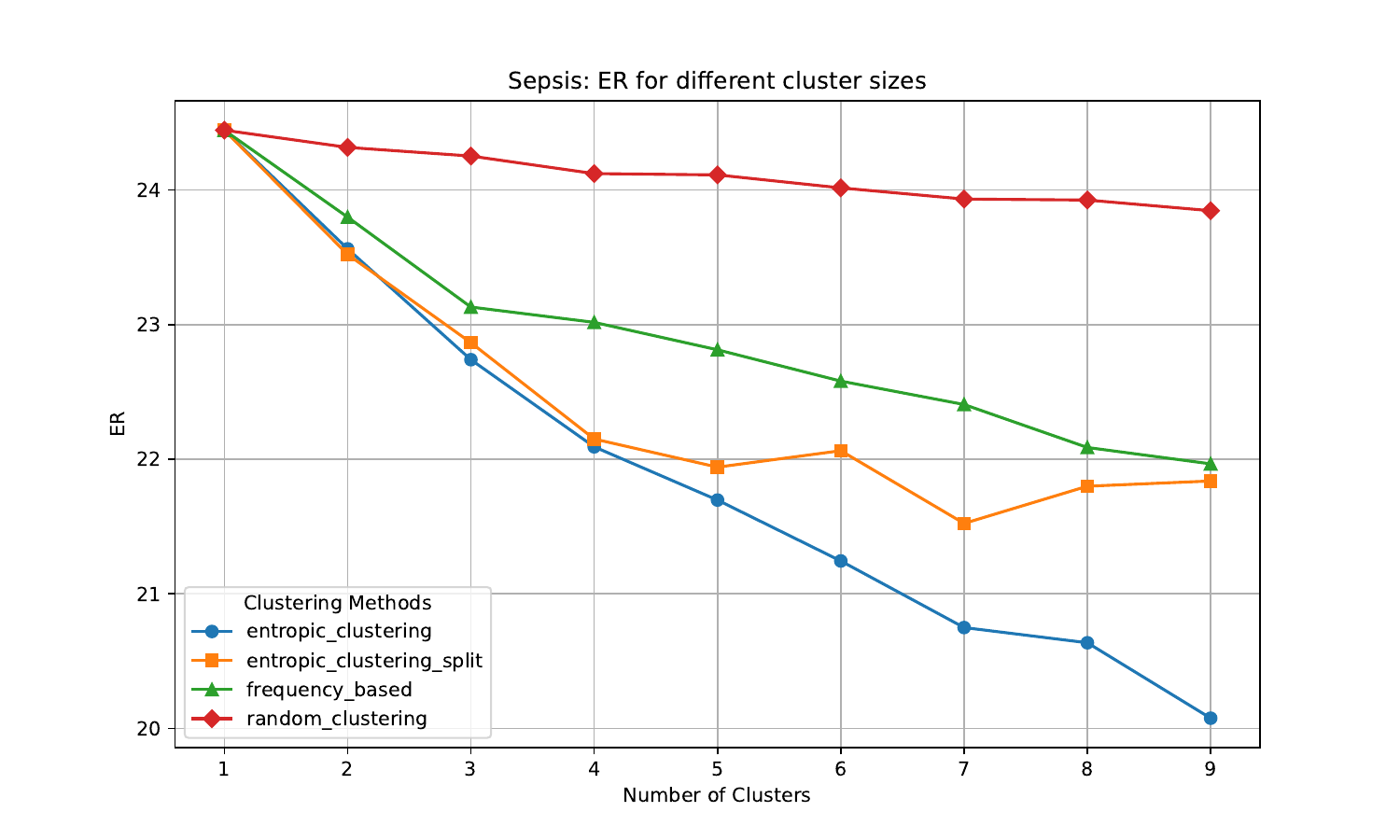}
    \end{subfigure}
    
    \vspace{0.5cm}
    
    \begin{subfigure}[b]{0.49\textwidth}
        \centering
        \includegraphics[trim={1.5cm 0.6cm 2.5cm 1.2cm}, clip, width=\linewidth]{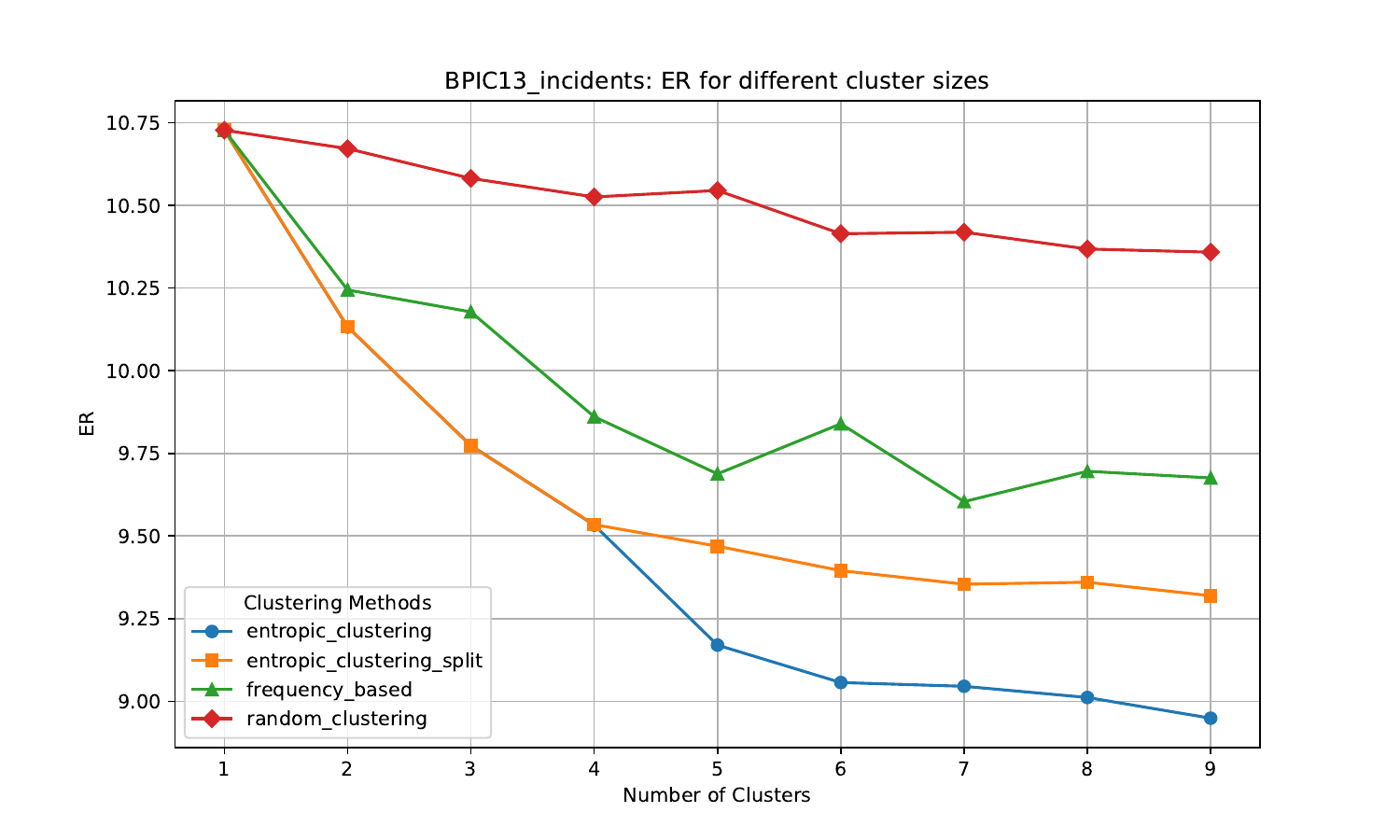}
    \end{subfigure}
    \hfill
    \begin{subfigure}[b]{0.49\textwidth}
        \centering
        \includegraphics[trim={1.5cm 0.6cm 2.5cm 1.2cm}, clip, width=\linewidth]{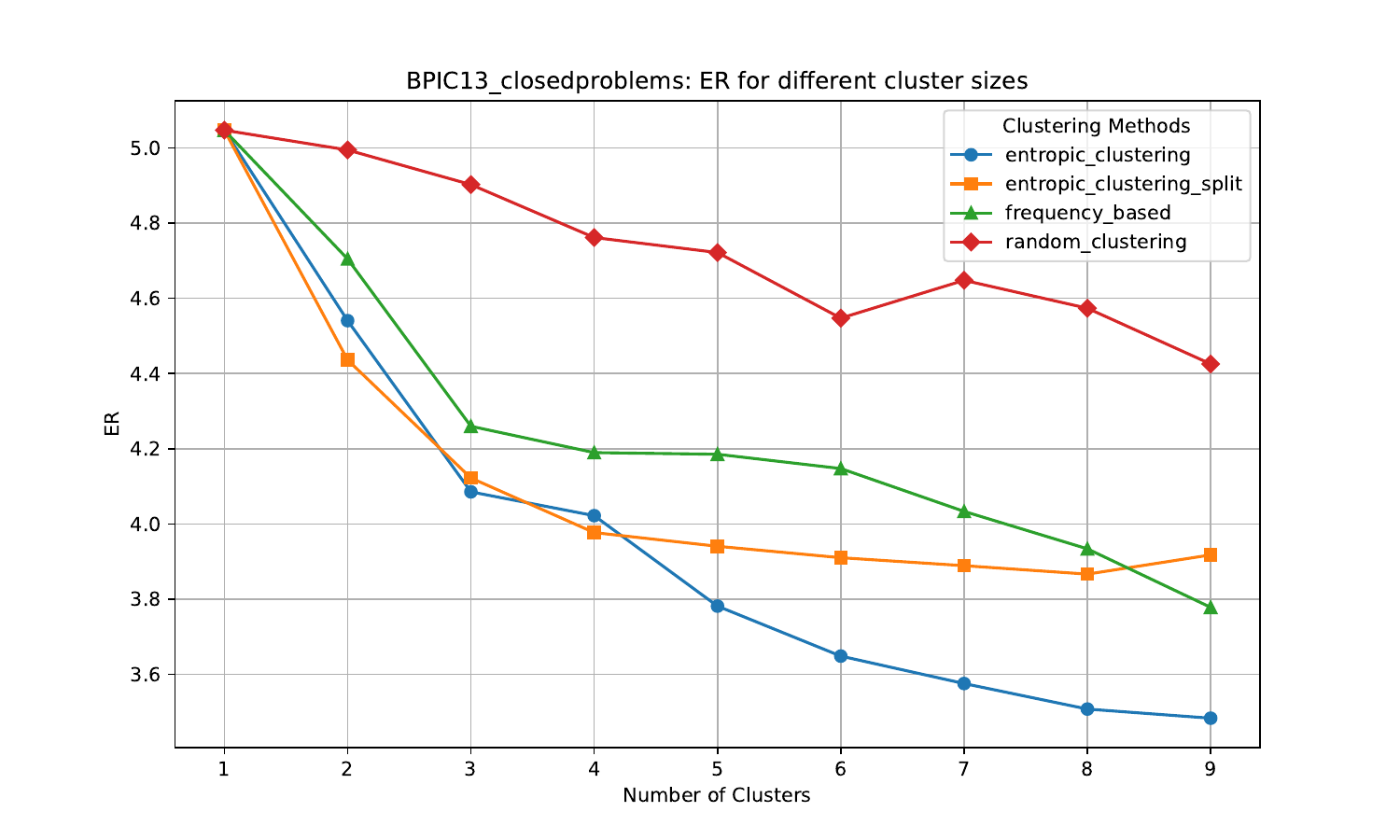}
    \end{subfigure}
    
    \vspace{0.5cm}
    
    \begin{subfigure}[b]{0.49\textwidth}
        \centering
        \includegraphics[trim={1.5cm 0.6cm 2.5cm 1.2cm}, clip, width=\linewidth]{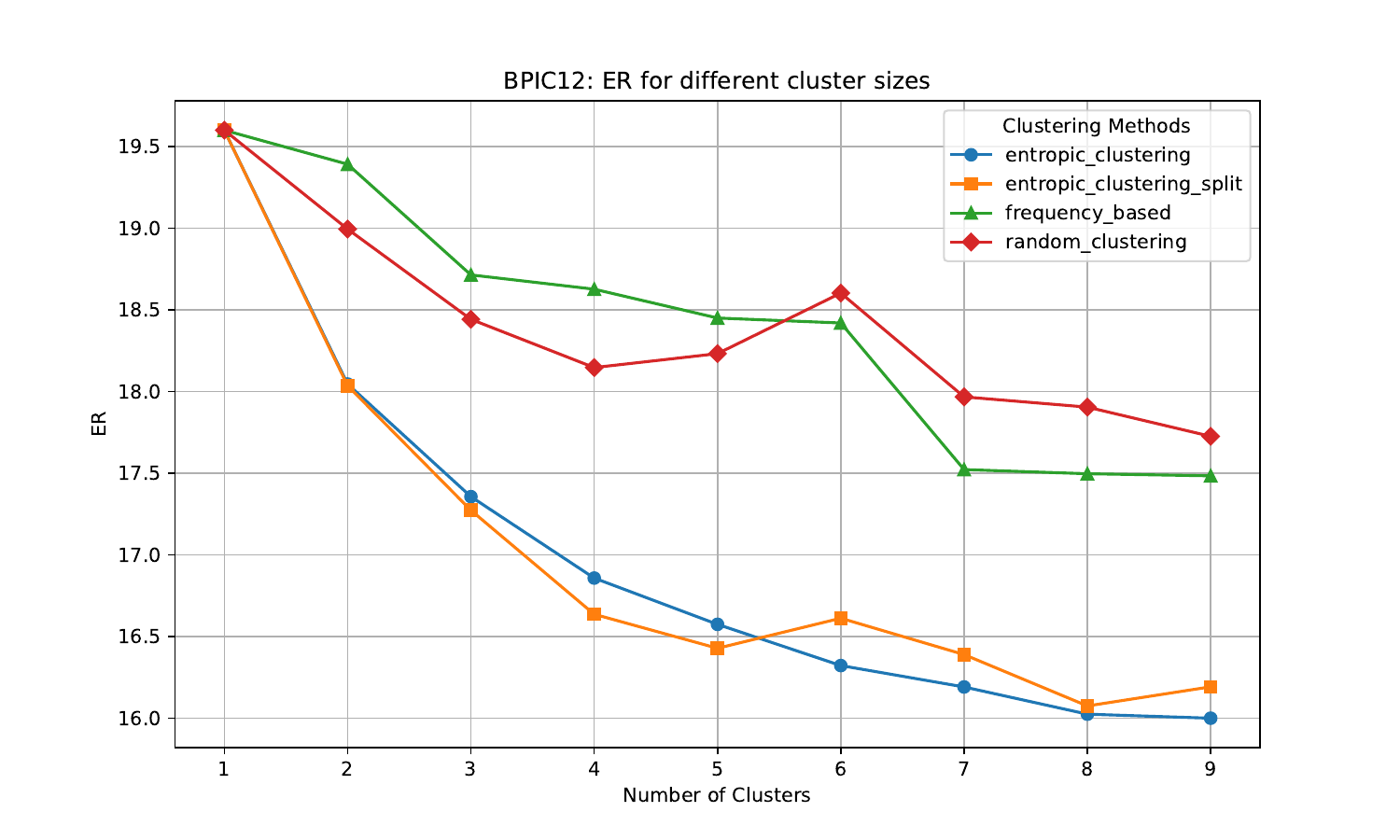}
    \end{subfigure}
    \hfill
    \begin{subfigure}[b]{0.49\textwidth}
        \centering
        \includegraphics[trim={1.5cm 0.6cm 2.5cm 1.2cm}, clip, width=\linewidth]{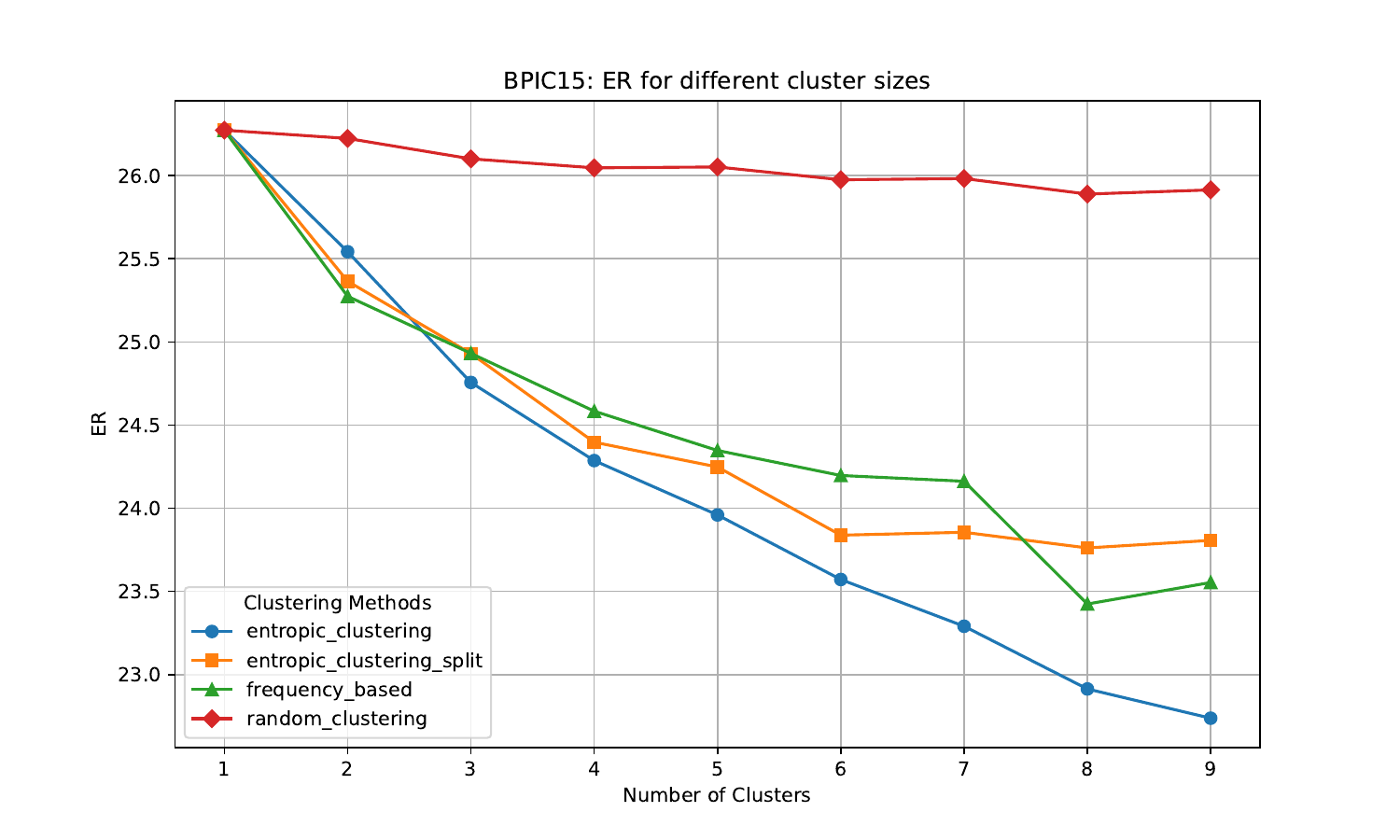}
    \end{subfigure}
    
    \caption{The elbow experiment with (simplified) ER, averaged per trace.}
    \label{fig:ER_elbow}
\end{figure}

\begin{figure}[htbp]
    \centering
    \begin{subfigure}[b]{0.49\textwidth}
        \centering
        \includegraphics[trim={1.5cm 0.6cm 2.5cm 1.2cm}, clip, width=\linewidth]{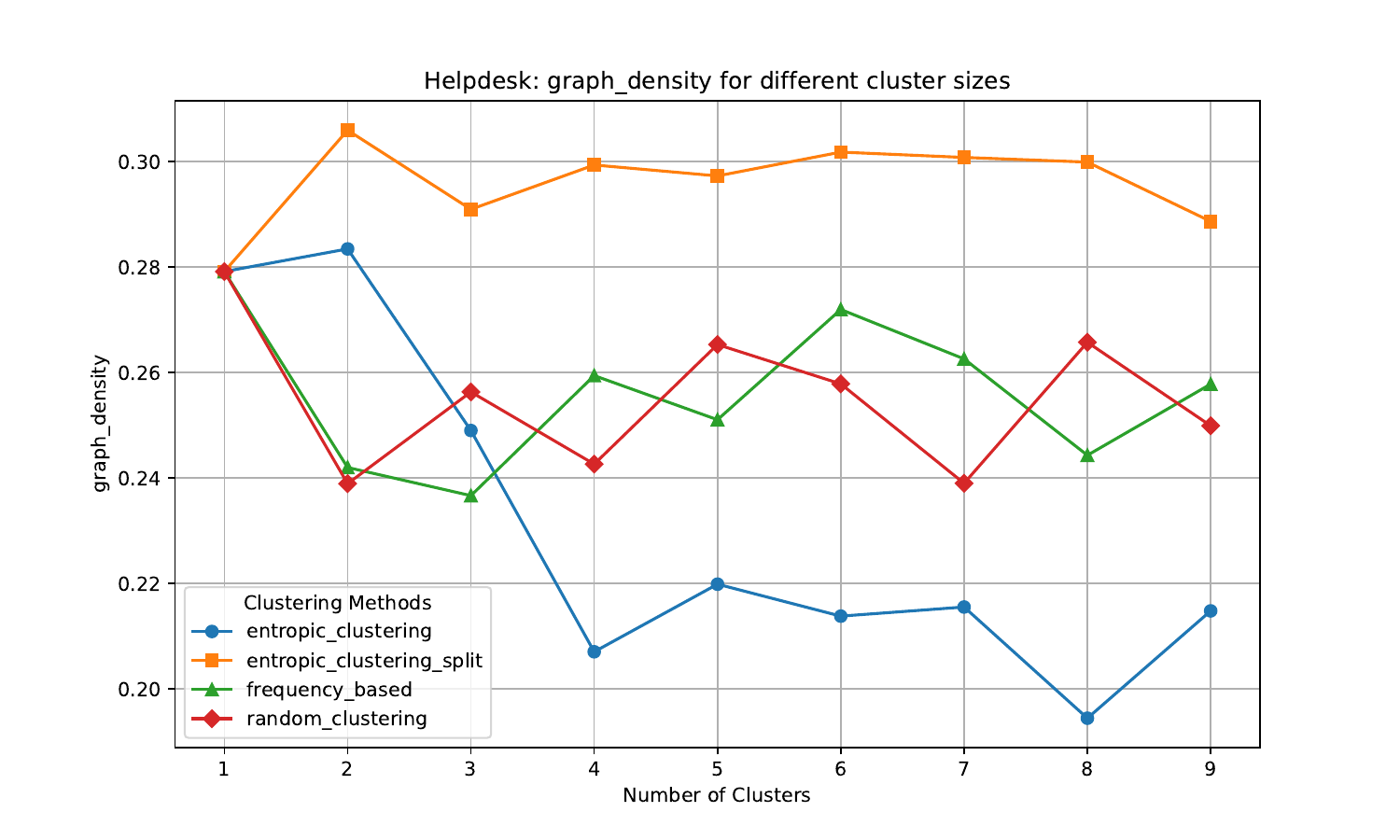}
    \end{subfigure}
    \hfill
    \begin{subfigure}[b]{0.49\textwidth}
        \centering
        \includegraphics[trim={1.5cm 0.6cm 2.5cm 1.2cm}, clip, width=\linewidth]{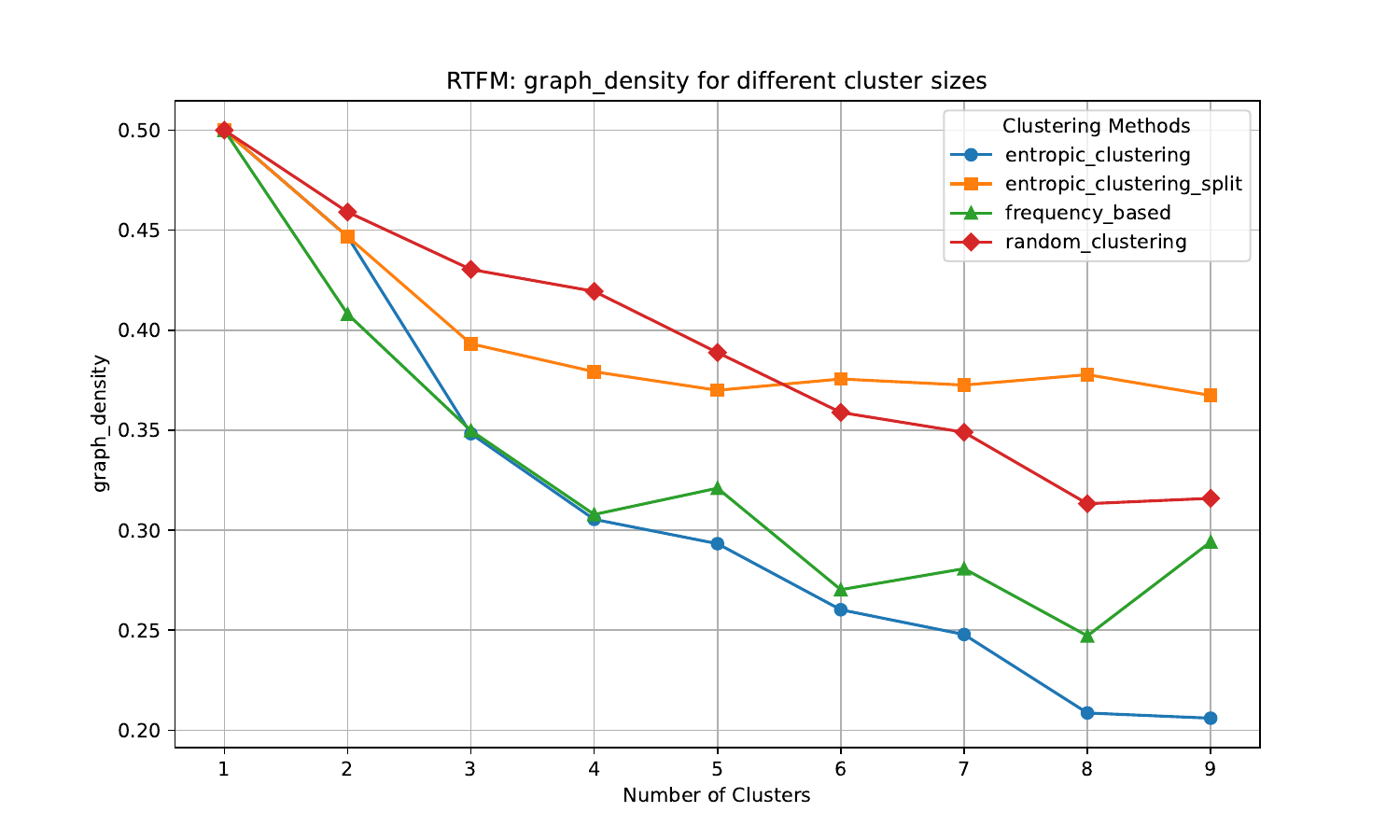}
    \end{subfigure}
    
    \vspace{0.5cm} 
    
    \begin{subfigure}[b]{0.49\textwidth}
        \centering
        \includegraphics[trim={1.5cm 0.6cm 2.5cm 1.2cm}, clip, width=\linewidth]{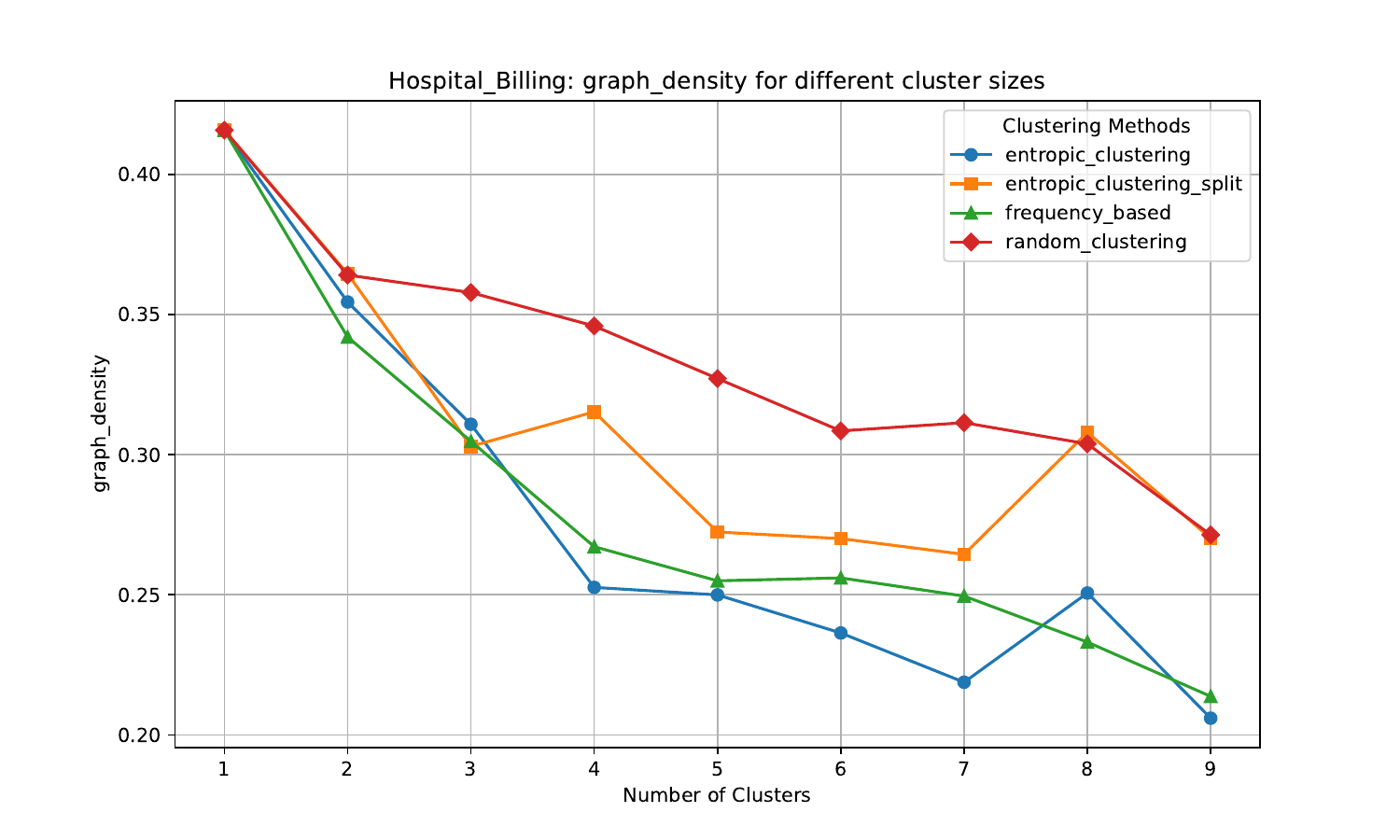}
    \end{subfigure}
    \hfill
    \begin{subfigure}[b]{0.49\textwidth}
        \centering
        \includegraphics[trim={1.5cm 0.6cm 2.5cm 1.2cm}, clip, width=\linewidth]{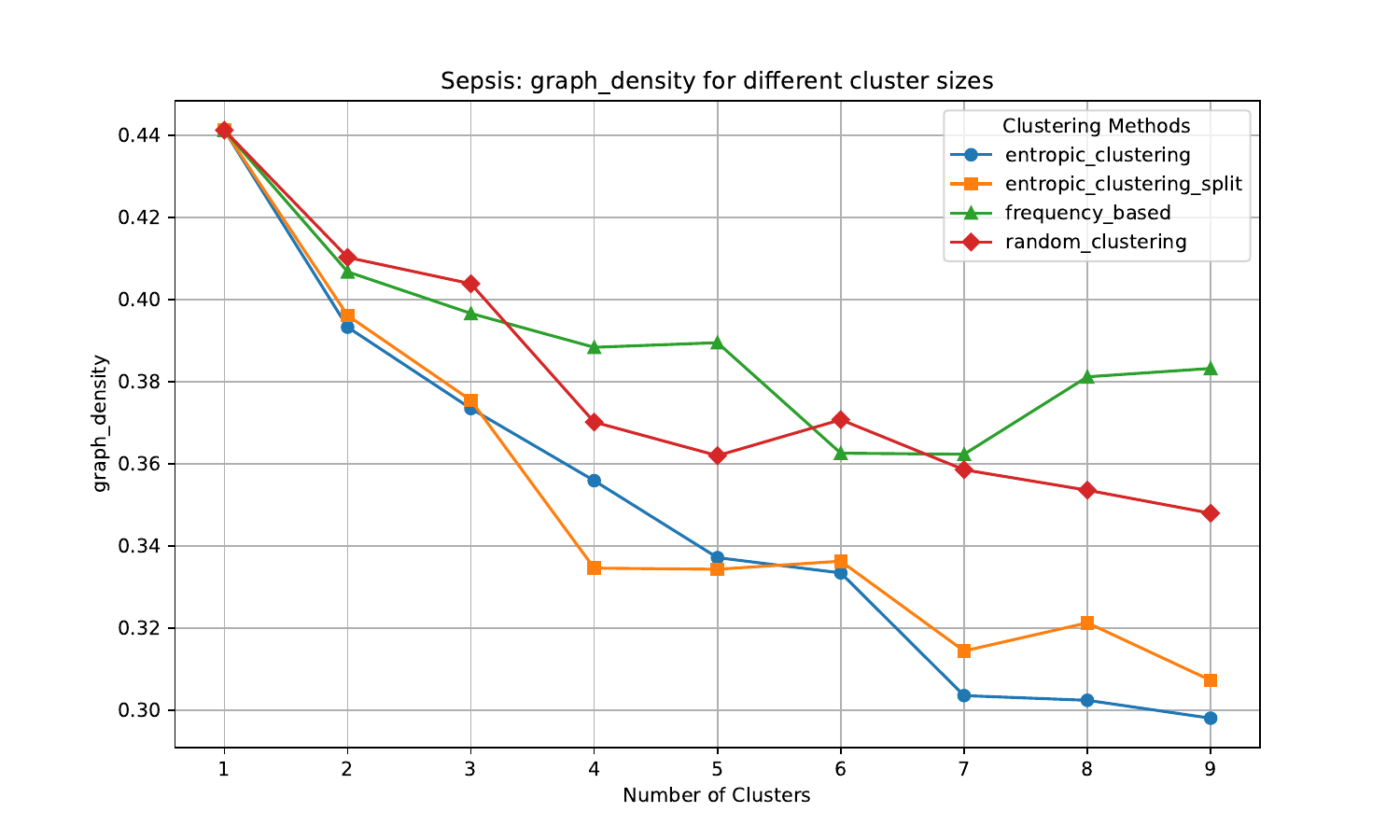}
    \end{subfigure}
    
    \vspace{0.5cm}
    
    \begin{subfigure}[b]{0.49\textwidth}
        \centering
        \includegraphics[trim={1.5cm 0.6cm 2.5cm 1.2cm}, clip, width=\linewidth]{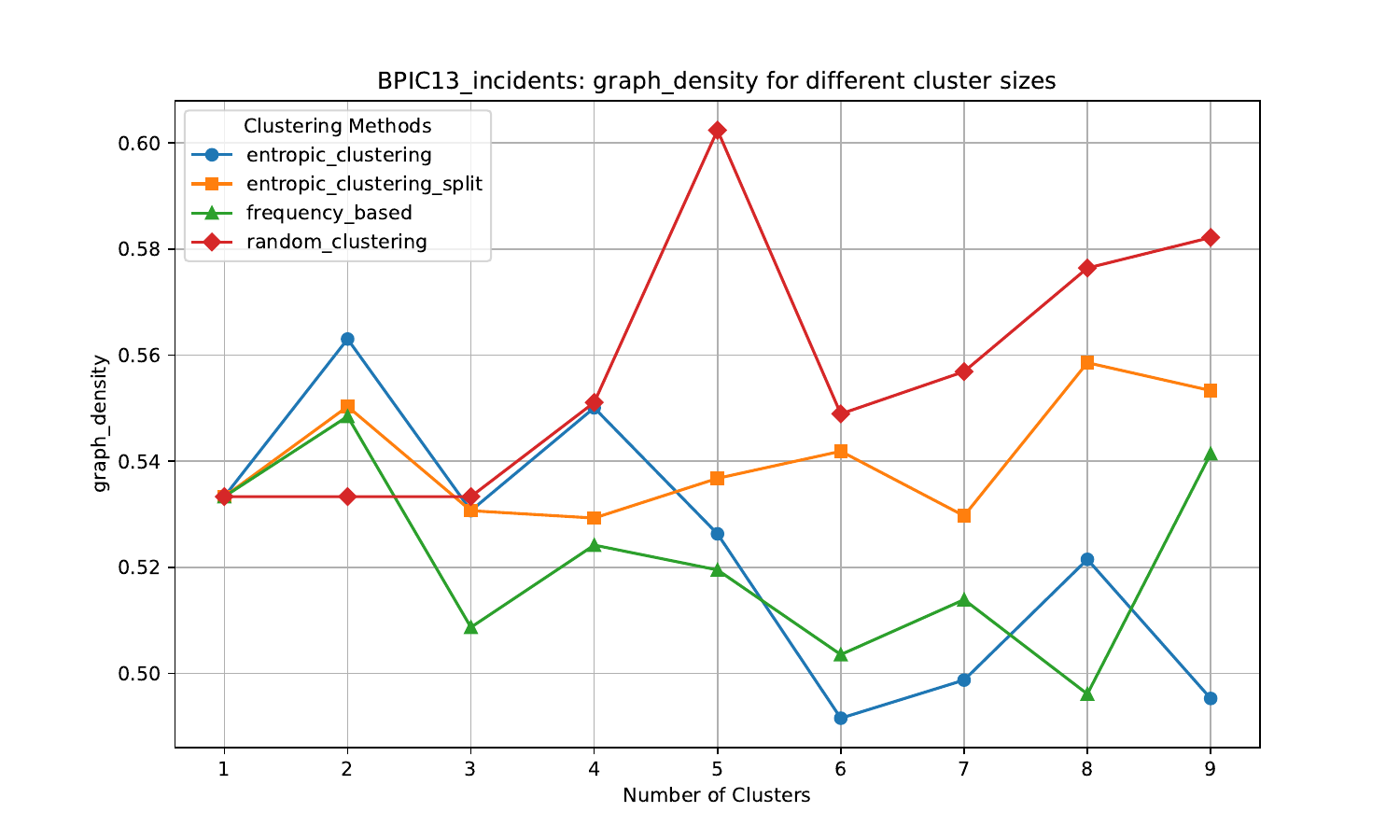}
    \end{subfigure}
    \hfill
    \begin{subfigure}[b]{0.49\textwidth}
        \centering
        \includegraphics[trim={1.5cm 0.6cm 2.5cm 1.2cm}, clip, width=\linewidth]{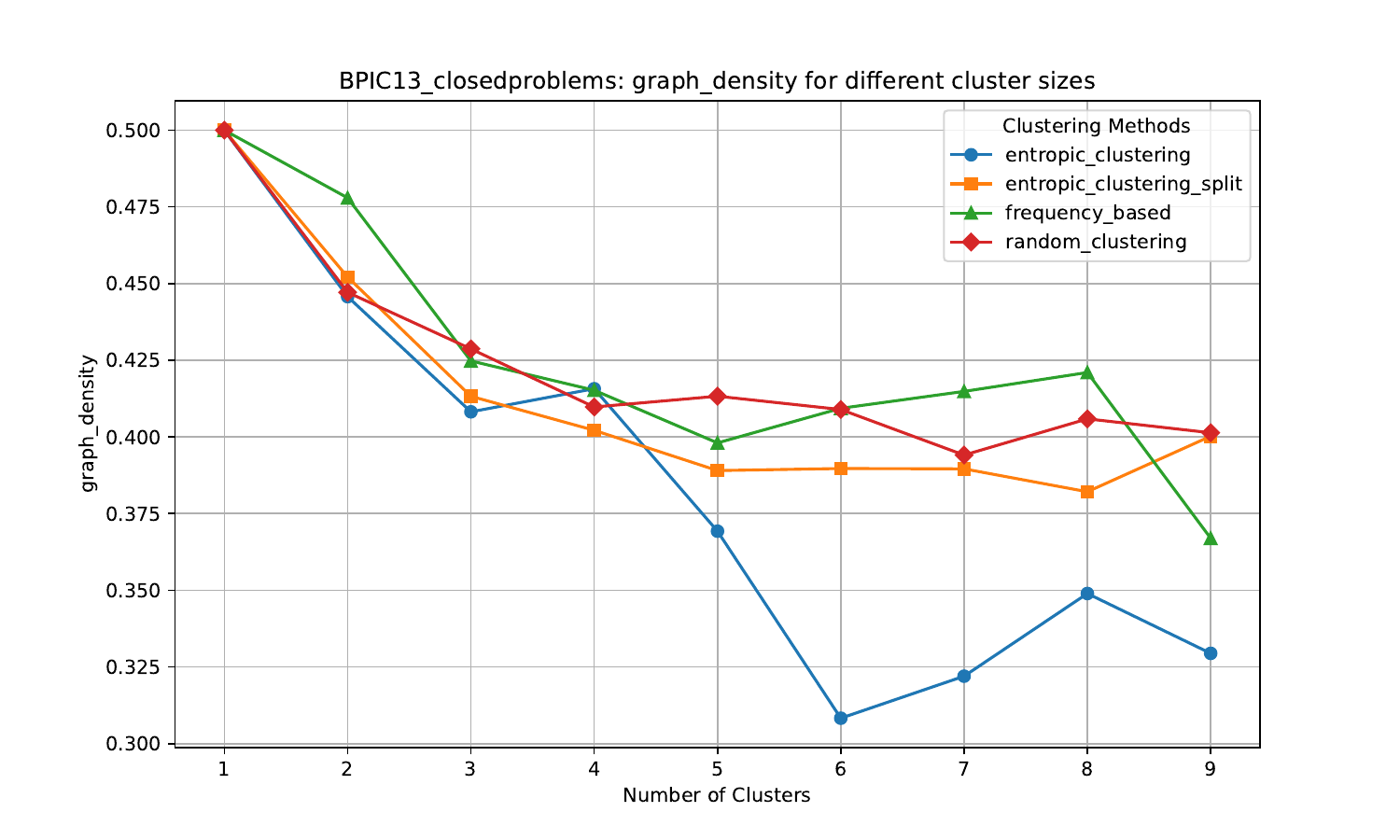}
    \end{subfigure}
    
    \vspace{0.5cm}
    
    \begin{subfigure}[b]{0.49\textwidth}
        \centering
        \includegraphics[trim={1.5cm 0.6cm 2.5cm 1.2cm}, clip, width=\linewidth]{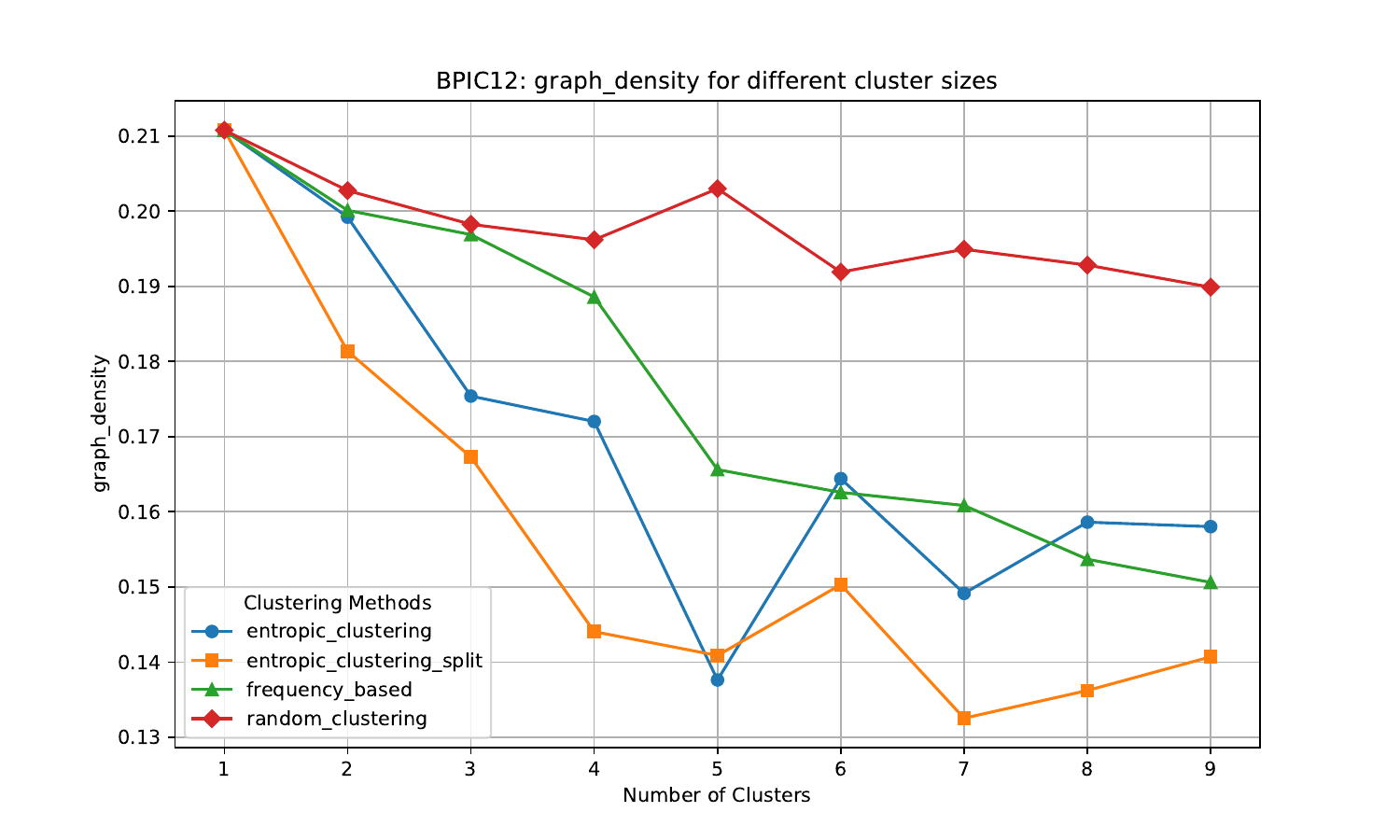}
    \end{subfigure}
    \hfill
    \begin{subfigure}[b]{0.49\textwidth}
        \centering
        \includegraphics[trim={1.5cm 0.6cm 2.5cm 1.2cm}, clip, width=\linewidth]{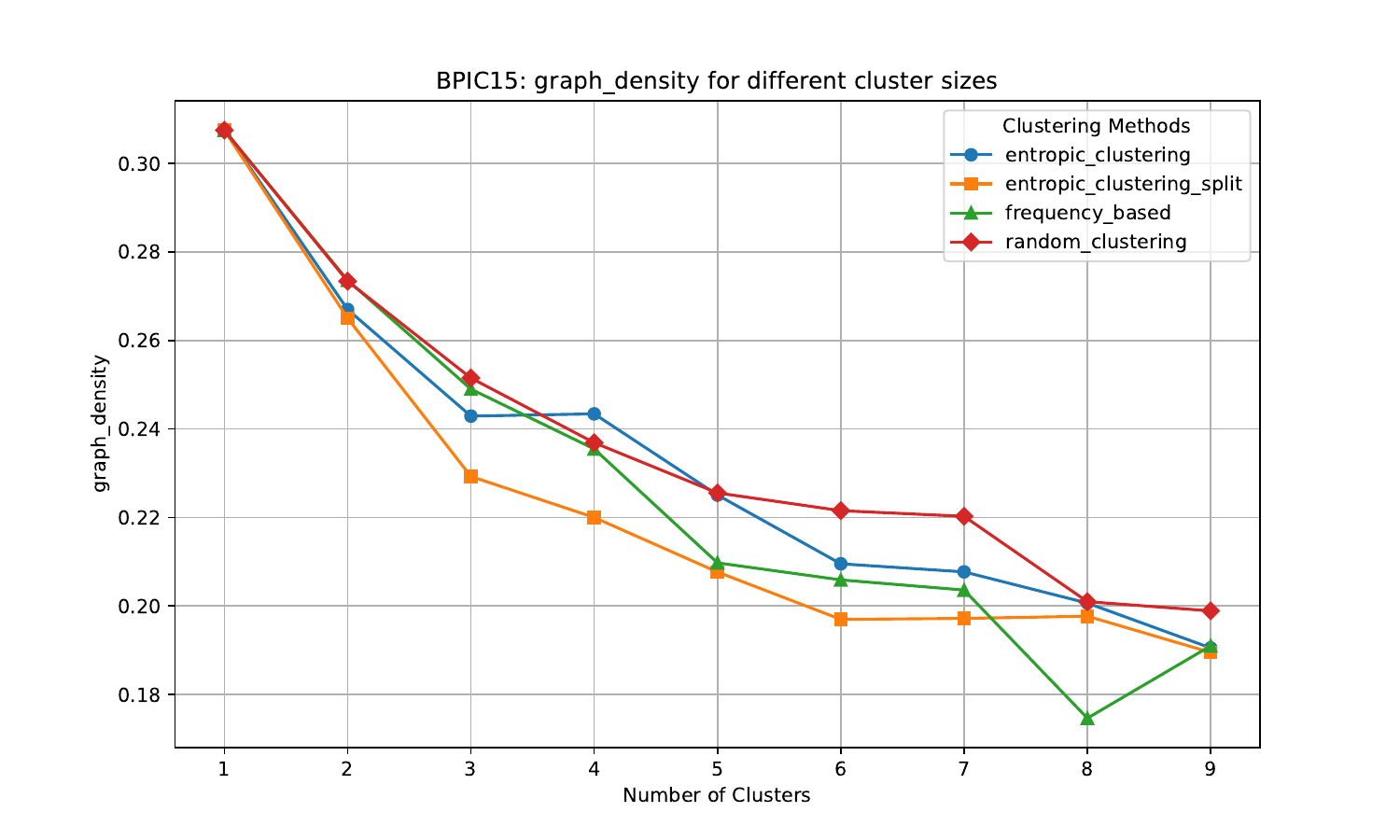}
    \end{subfigure}
    
    \caption{The elbow experiment with graph density.}
    \label{fig:graph_density_elbow}
\end{figure}

\begin{figure}[htbp]
    \centering
    \begin{subfigure}[b]{0.49\textwidth}
        \centering
        \includegraphics[trim={1.5cm 0.6cm 2.5cm 1.2cm}, clip, width=\linewidth]{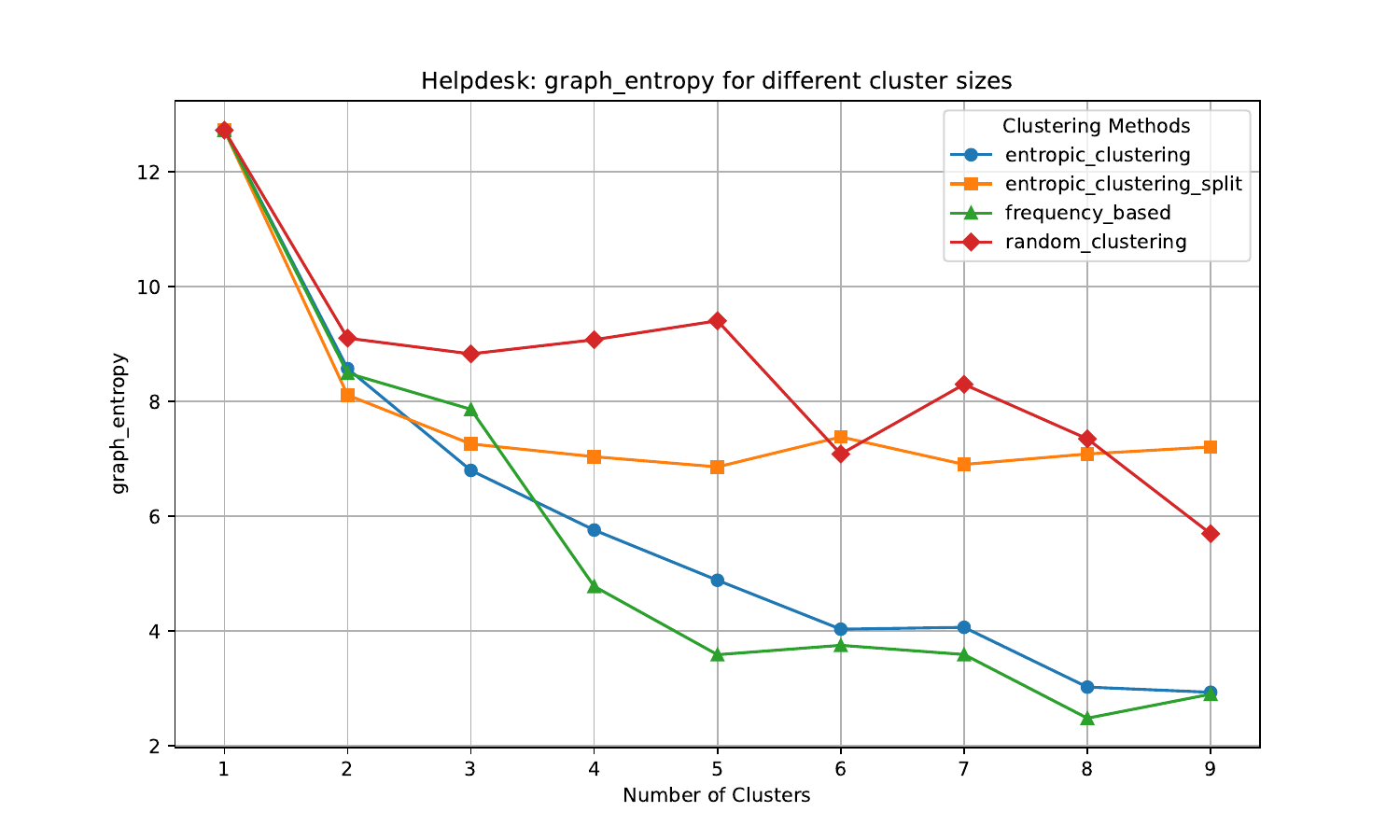}
    \end{subfigure}
    \hfill
    \begin{subfigure}[b]{0.49\textwidth}
        \centering
        \includegraphics[trim={1.5cm 0.6cm 2.5cm 1.2cm}, clip, width=\linewidth]{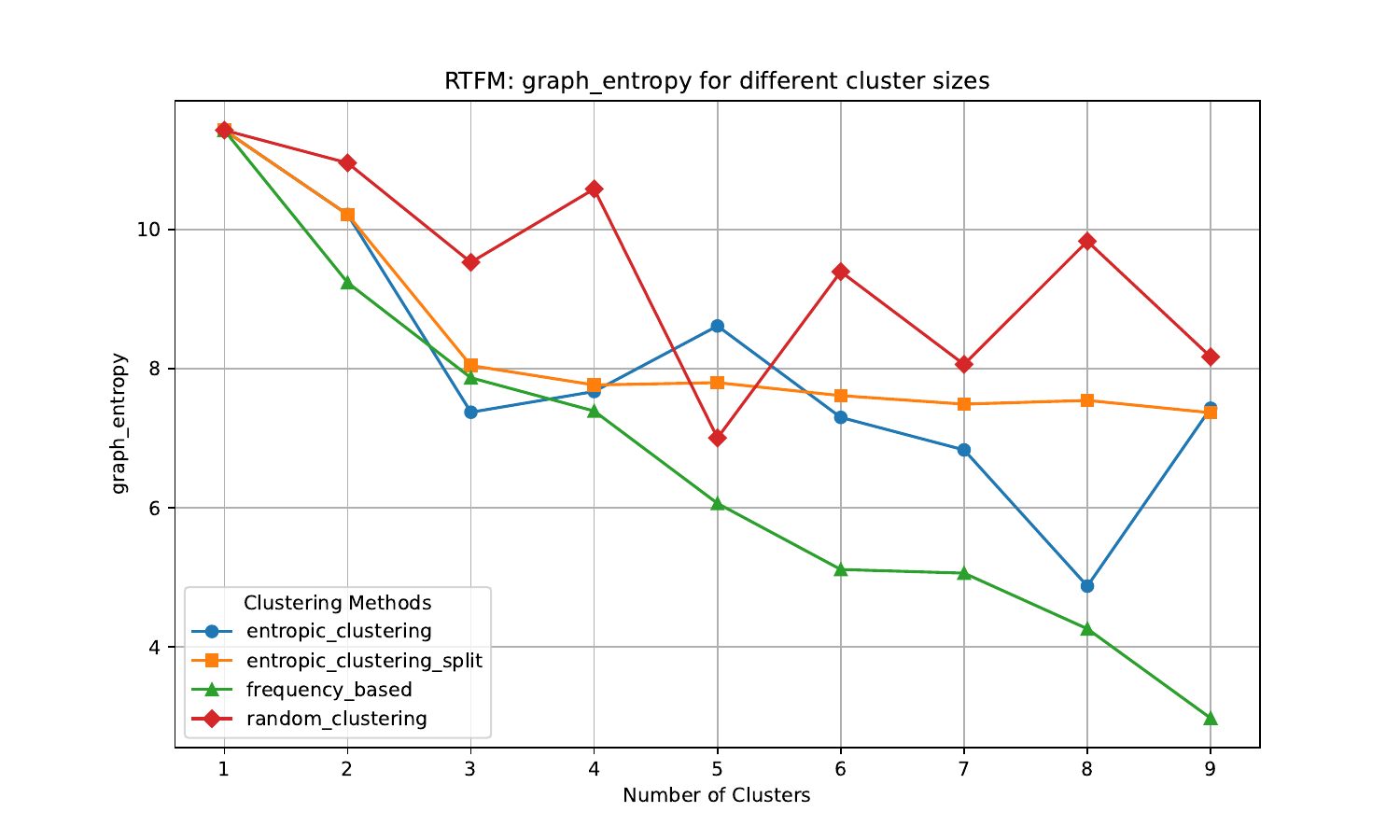}
    \end{subfigure}
    
    \vspace{0.5cm} 
    
    \begin{subfigure}[b]{0.49\textwidth}
        \centering
        \includegraphics[trim={1.5cm 0.6cm 2.5cm 1.2cm}, clip, width=\linewidth]{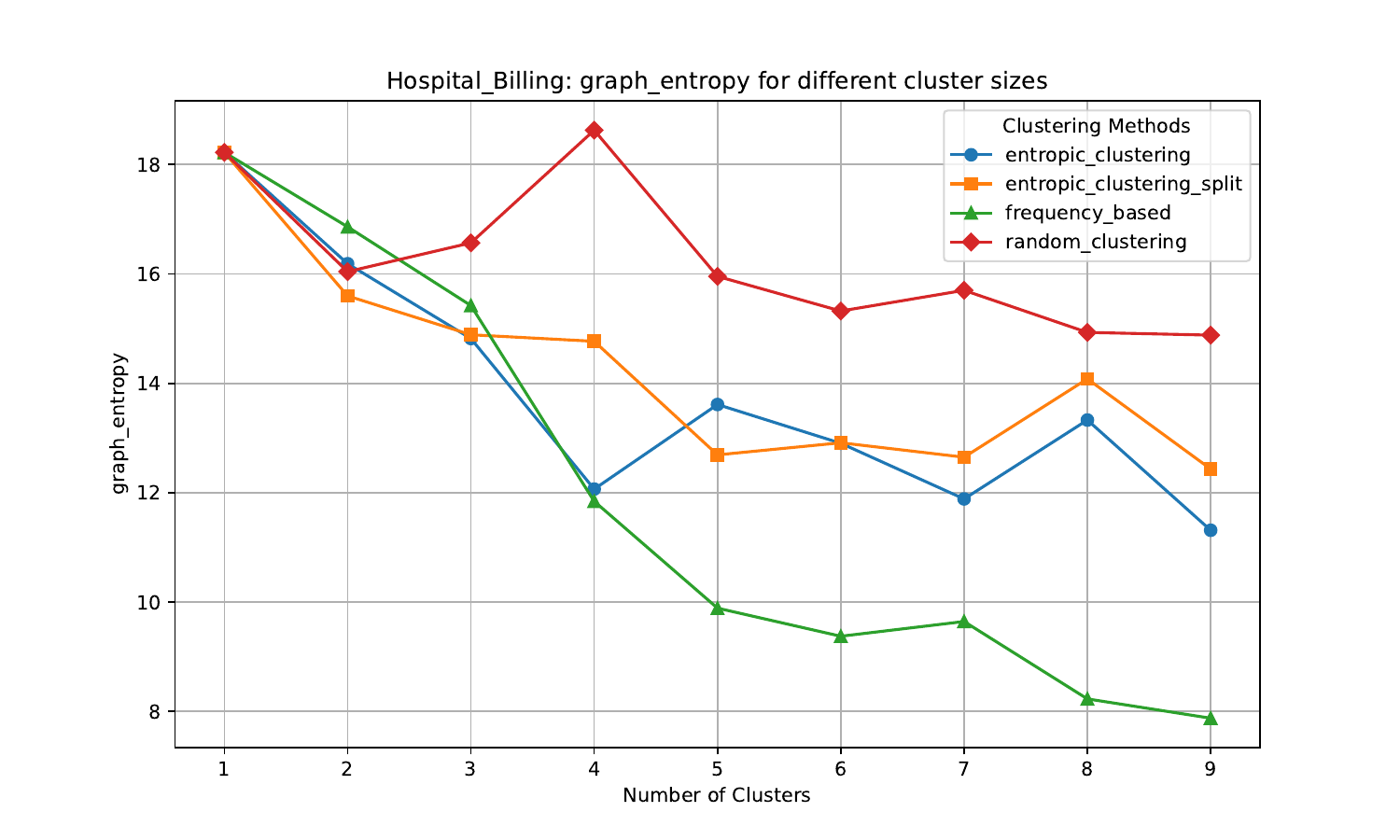}
    \end{subfigure}
    \hfill
    \begin{subfigure}[b]{0.49\textwidth}
        \centering
        \includegraphics[trim={1.5cm 0.6cm 2.5cm 1.2cm}, clip, width=\linewidth]{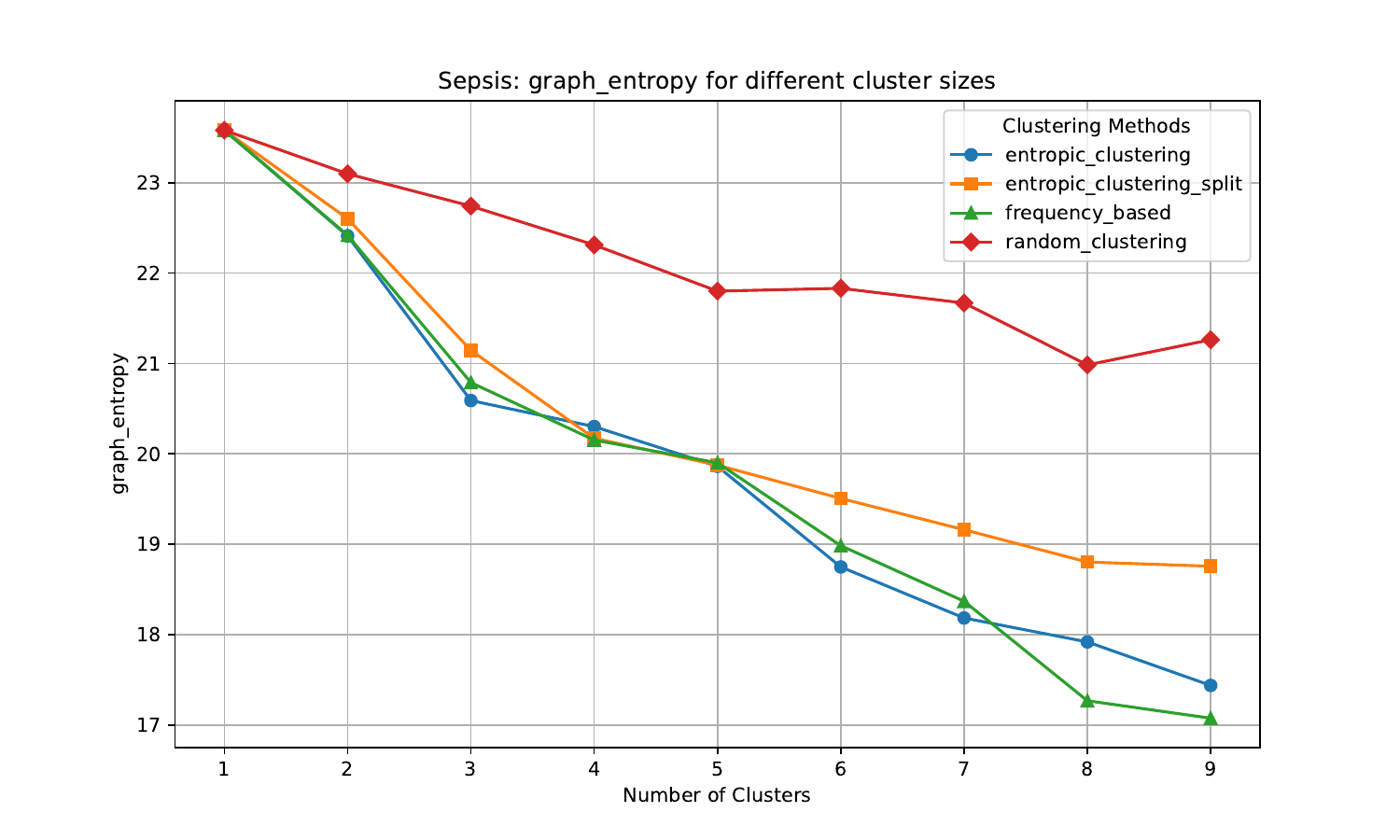}
    \end{subfigure}
    
    \vspace{0.5cm}
    
    \begin{subfigure}[b]{0.49\textwidth}
        \centering
        \includegraphics[trim={1.5cm 0.6cm 2.5cm 1.2cm}, clip, width=\linewidth]{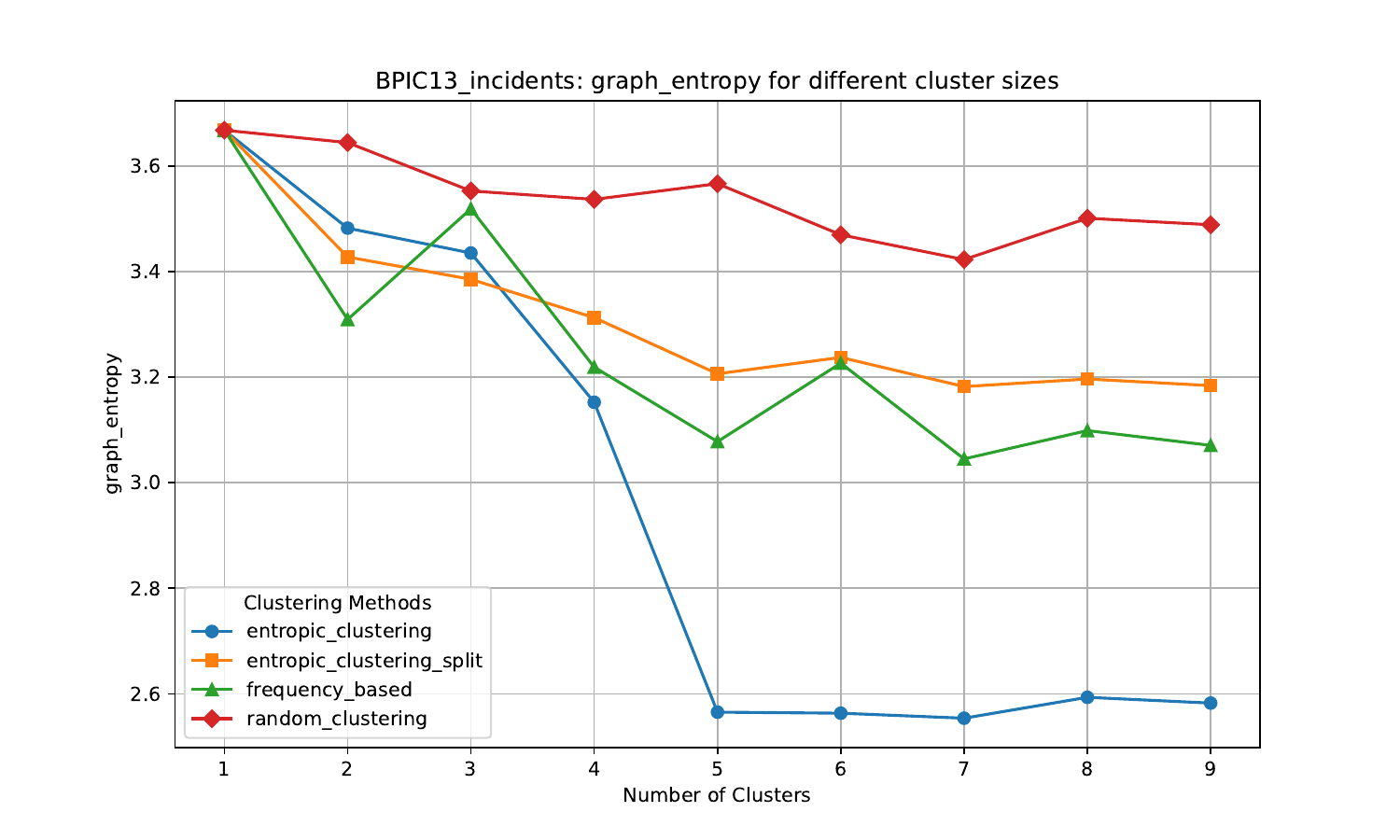}
    \end{subfigure}
    \hfill
    \begin{subfigure}[b]{0.49\textwidth}
        \centering
        \includegraphics[trim={1.5cm 0.6cm 2.5cm 1.2cm}, clip, width=\linewidth]{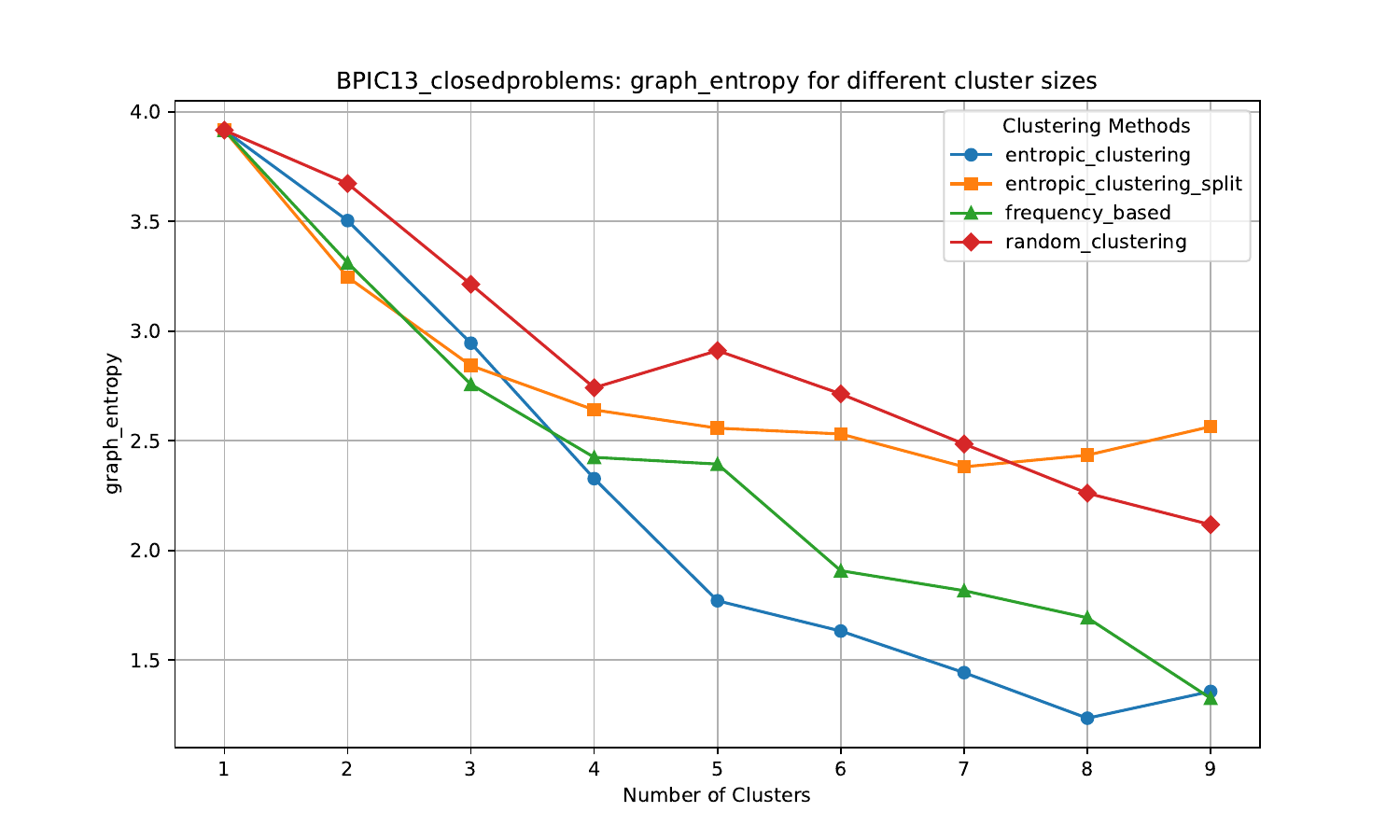}
    \end{subfigure}
    
    \vspace{0.5cm}
    
    \begin{subfigure}[b]{0.49\textwidth}
        \centering
        \includegraphics[trim={1.5cm 0.6cm 2.5cm 1.2cm}, clip, width=\linewidth]{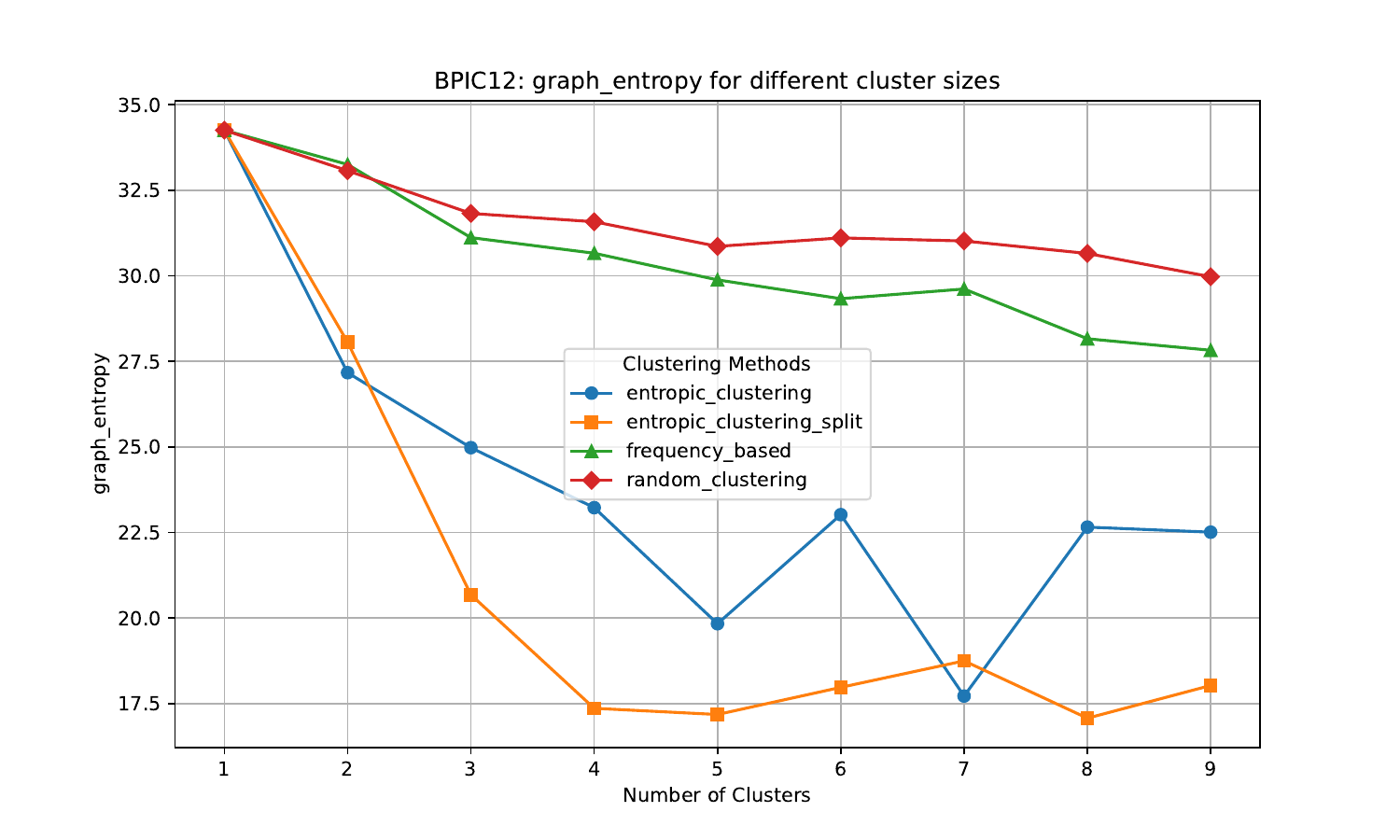}
    \end{subfigure}
    \hfill
    \begin{subfigure}[b]{0.49\textwidth}
        \centering
        \includegraphics[trim={1.5cm 0.6cm 2.5cm 1.2cm}, clip, width=\linewidth]{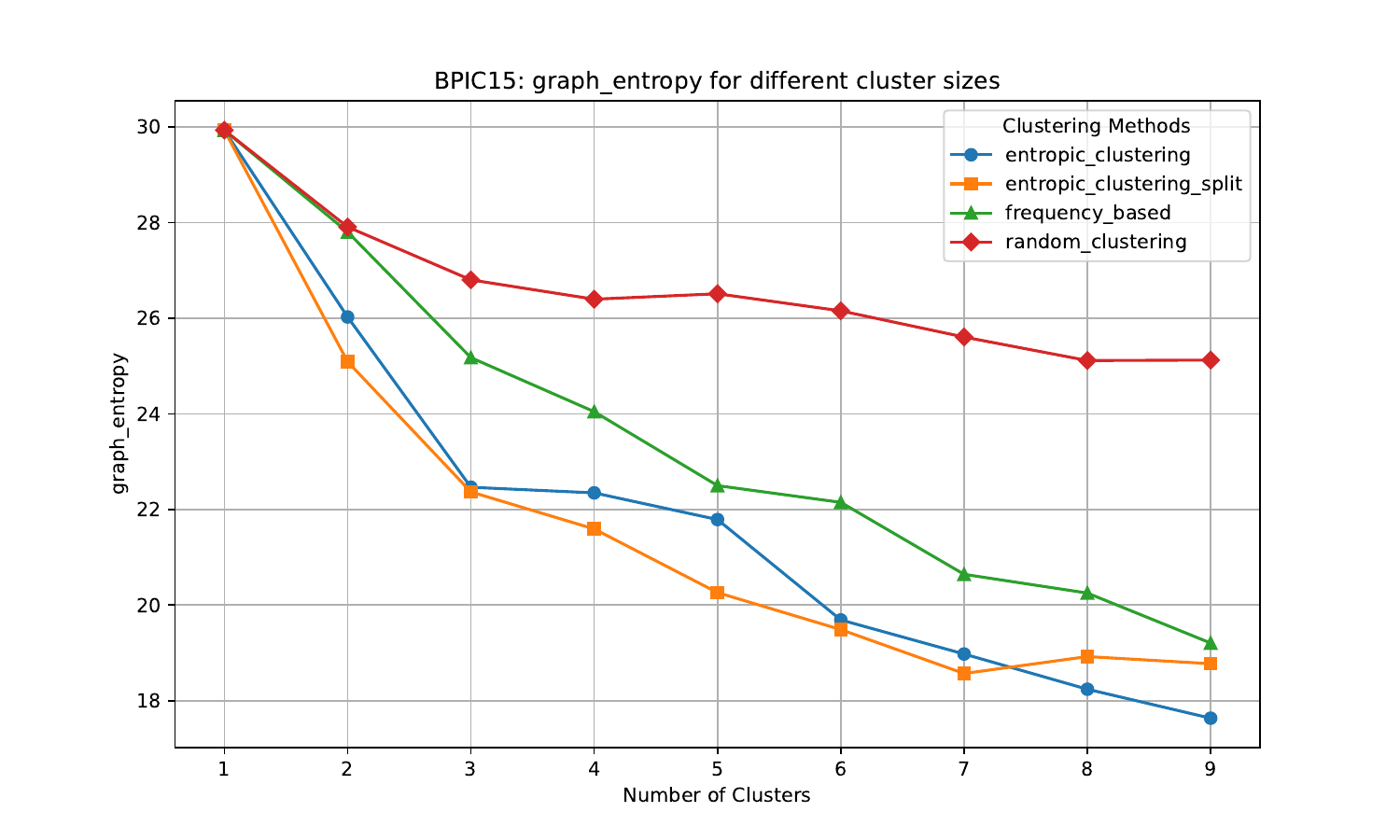}
    \end{subfigure}
    
    \caption{The elbow experiment with graph entropy.}
    \label{fig:graph_entropy_elbow}
\end{figure}

This appendix presents the detailed results of the parameter determination phase. To select an appropriate number of clusters ($k$) for each event log, we analyzed the evolution of model quality metrics across a range of cluster counts, from $k=1$ (the original unclustered log) to $k=10$. For each $k$, we executed four clustering variants: random clustering, frequency-based clustering (with kmeans++), \texttt{EC}, and \texttt{EC\textsubscript{split}} (both utilizing \texttt{++} initialization). To ensure a representative evaluation of the resulting partition, the metric scores reported in the figures below are calculated as the weighted average over the generated clusters, where the weight corresponds to the number of traces assigned to each cluster. The selection of $k$ was based on the ``elbow method'', identifying the point where the marginal improvement in the metrics (e.g., reduction in Entropic Relevance) diminishes significantly. Figure~\ref{fig:ER_elbow} displays the ER (average per trace) scores for different numbers of clusters, Figure~\ref{fig:graph_density_elbow} the graph density, and Figure~\ref{fig:graph_entropy_elbow} the graph entropy. Scores for $k = 1$ correspond to no clustering, i.e. calculated on the whole event log.

\end{document}